\documentclass[10pt,twocolumn,letterpaper]{article}

\usepackage[pagenumbers]{cvpr} 

\usepackage{array,booktabs,graphicx}
\newcolumntype{C}[1]{>{\centering\arraybackslash}p{#1}}
\usepackage{booktabs,multirow,xcolor,pifont,graphicx,array}
\usepackage{xfrac}  

\definecolor{cvprblue}{rgb}{0.21,0.49,0.74}
\usepackage[pagebackref,breaklinks,colorlinks,allcolors=cvprblue]{hyperref}
\usepackage[table,xcdraw]{xcolor}
\def\paperID{4642} 
\def\confName{CVPR}
\def\confYear{2026}
\newcommand{\name}{CALMARS}
\title{
\raisebox{-0.2\height}{\includegraphics[width=0.8cm]{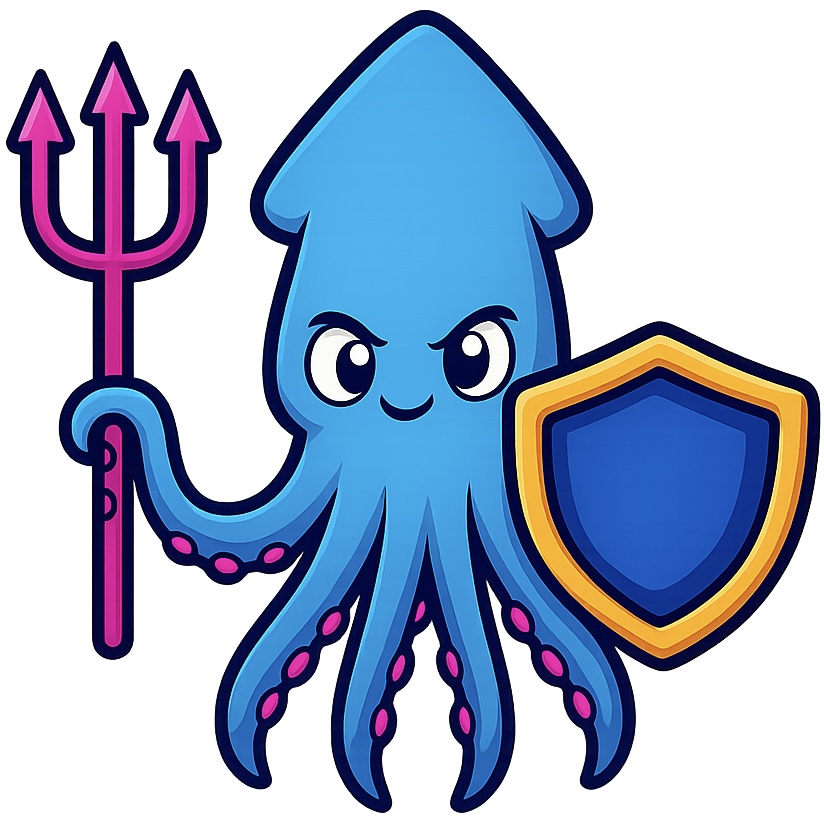}}\hspace{0.3em}
\textbf{CALMARS: Towards Advanced and Robust Unified Multi-modal Encoders via Multi-stage Adversarial Training}
}

\author{
Chih-Ting Liao$^{1}$\quad
Zhangquan Chen$^{2}$\quad
Chunlei Meng$^{3}$\quad
Tzu-Yu Huang$^{4}$\quad
Xin Cao$^{1}$\quad
Xu Zheng$^{5,*}$\\[3pt]
$^{1}$UNSW Sydney\quad 
$^{2}$Tsinghua University\quad 
$^{3}$Fudan University\quad
$^{4}$UTS\quad 
$^{5}$HKUST(GZ)\\
{\tt\small zhengxu128@gmail.com} \\[3pt]
$^{*}$Corresponding author
}

\begin{document}
\maketitle
\begin{abstract}
Endeavors have been made in learning unified multi-modal encoders while the robustness under adversarial perturbations remains unexplored—a critical concern for safety-sensitive applications. 
In this paper, we investigate this problem and find that even mild perturbations lead to substantial performance drops, while non-visual inputs (e.g. audio and point cloud) are especially vulnerable, often failing entirely.
To this end, we propose \textbf{\name}, the first efficient multi-stage adversarial training framework with Clean-oriented Alignment and Latent Modeling (CALM) and Multi-phase Adversarial Representation Stabilization (MARS) modules, aiming at improving robustness across six modalities. \name~keeps pretrained encoders and semantic centers frozen, aiming at ensuring compatibility with existing foundation models while preserving unified embedding space. Specifically, CALM focuses on improving clean example performance with feature distillation and embedding calibration while MARS is designed to enhance adversarial robustness with multi-phase fine-tuning.
Moreover, existing targeted attacks fails in the overall ranking qualify, such as high Top-5 results. Thus we propose \textbf{CrossMaxim}, a non-targeted attack that acts like a "machine-gun sweep",repelling representations from their ground truth without relying on external hard-negatives.
Experiments across six modalities and three Bind models demonstrate that \name achieves state-of-the-art performance, while preserving or even \textbf{improving} clean zero-shot and retrieval performance with less than \textbf{\textit{1\%}} trainable parameters.
\end{abstract}
    
\section{Introduction}
\label{sec:intro}

Modern multi-modal foundation models (e.g., ImageBind, LanguageBind, UniBind) aim to unify diverse sensory modalities—including image, audio, video, point cloud, thermal, and event streams—within a shared semantic space, enabling flexible cross-modal reasoning and retrieval \cite{girdhar2023imagebind,zhu2023languagebind,lyu2024unibind}. 
While these models achieve impressive zero-shot and transfer performance under clean conditions, their \textit{robustness to adversarial perturbations remains largely unexplored} \cite{zhou2024revisiting}. 
This limitation poses serious risks for safety-critical applications such as autonomous driving, embodied AI, and healthcare, where small perturbations in a single modality can propagate through the shared embedding space and cause cascading failures across modalities.

Prior robustness studies on vision–language models have focused mainly on CLIP, revealing vulnerability to both distribution shifts and adversarial examples \cite{fang2022data,nguyen2022quality,crabbe2024interpreting,goodfellow2015explaining}. 
Classical defenses such as TRADES, TeCoA, and LAAT introduce contrastive or margin-oriented objectives to stabilize visual–textual alignment \cite{zhang2019theoretically,mao2023tecoa,li2024language}, while FARE (RobustCLIP) proposes an unsupervised adversarial fine-tuning strategy \cite{schlarmann2024robust}. 
However, these methods are largely restricted to image–text pairs, typically require full-encoder tuning, and do not directly generalize to foundation models with frozen multi-modal backbones. 
Moreover, unified handling of non-visual modalities is still missing; robustness degradation can be especially severe—e.g., UniBind exhibits near-collapse on audio and point-cloud under small perturbations \cite{lyu2024unibind}.

To address these challenges, we present \textbf{CALMARS}, a unified and efficient \textit{multi-stage adversarial training framework} for Bind-style multi-modal encoders. 
CALMARS comprises two complementary stages: 
(1) \textit{Clean-oriented Alignment and Latent Modeling (CALM)} improves clean performance through feature distillation and embedding calibration without modifying the encoders \cite{yang2024clip,csizmadia2025distill}; and 
(2) \textit{Multi-phase Adversarial Representation Stabilization (MARS)} progressively enhances robustness via geometric consistency, distributional regularization, and counterfactual margin separation, aligning with recent advances in multimodal adversarial training \cite{waseda2024mat}. 
Both stages operate on lightweight, modality-specific projection heads (\textless1\% parameters), preserving the pretrained semantic space and ensuring plug-and-play compatibility with existing models.

Furthermore, we introduce \textbf{CrossMaxim}, an untargeted retrieval attack that perturbs the entire similarity space—breaking not only top-1 but also top-5 alignment—thereby exposing vulnerabilities that targeted attacks (e.g., CrossFire) may miss \cite{dou2023adversarial}. 
We follow reliable evaluation practices using diverse, strong attacks \cite{croce2020reliable}. 
This yields the first comprehensive multi-modal robustness benchmark across six modalities and multiple Bind-style architectures \cite{deng2009imagenet,zhou2017places,jia2021llvip,piczak2015esc,xu2016msr,soomro2012ucf101,lin2014microsoft,hodosh2013framing}. 

Overall, our contributions are summarized as follows: 
\textbf{(I)} We establish the first large-scale evaluation of adversarial robustness in unified multi-modal foundation models, covering six modalities and three representative architectures \cite{girdhar2023imagebind,zhu2023languagebind,lyu2024unibind}. 
\textbf{(II)} We propose a two-stage pipeline (CALM + MARS) that improves both clean and adversarial robustness while keeping encoders frozen \cite{waseda2024mat}. 
\textbf{(III)} We introduce \textit{CrossMaxim}, a new untargeted retrieval attack that reveals deeper alignment vulnerabilities across modalities, complementing targeted attacks like CrossFire \cite{dou2023adversarial}. 
\textbf{(IV)} Extensive experiments demonstrate that CALMARS achieves state-of-the-art robustness–accuracy trade-offs across classification, retrieval, and transfer tasks, while maintaining or improving clean performance.

\section{Related Works}
\label{sec:related_works}
\textbf{Adversarial Robustness in VLMs.}  
Adversarial robustness in VLMs, particularly CLIP~\citep{radford2021learning}, has been widely studied. 
Despite strong zero-shot performance, CLIP remains highly vulnerable to adversarial perturbations~\citep{mao2023tecoa}. 
Early work such as TRADES~\citep{zhang2019theoretically} introduced a theoretically principled framework balancing accuracy and robustness through adversarial training. 
FARE, proposed in RobustCLIP~\citep{schlarmann2024robust}, is the first unsupervised adversarial fine-tuning framework for CLIP, aligning clean and adversarial embeddings without labels. 
TeCoA~\citep{mao2023tecoa} adopts text-guided contrastive fine-tuning to enhance multimodal consistency, while LAAT~\citep{li2024language} leverages language-driven anchors for zero-shot adversarial robustness. 

\noindent \textbf{Bind-Style Models} aim to embed diverse sensory inputs into a shared semantic space~\cite{girdhar2023imagebind,zhu2024languagebind,bagdasaryan2024adversarial,zheng2023deep,zhou2024exact,zhou2024eventbind,zheng2024magic++,zheng2024learning}. ImageBind~\citep{girdhar2023imagebind} achieves cross-modal alignment using image-paired data. LanguageBind~\citep{zhu2024languagebind} aligns all modalities to a frozen text encoder with language annotations, while UniBind~\citep{lyu2024unibind} introduces modality-agnostic alignment centers via large language models. Though these models show strong zero-shot and retrieval performance, they have only been evaluated under clean conditions. Bagdasaryan et al.~\citep{bagdasaryan2024adversarial} identified ImageBind’s vulnerability to cross-modal attacks, but no prior work has explored adversarial robustness for LanguageBind or UniBind. 

\begin{figure}[t]
    \centering
    \includegraphics[width=0.9\linewidth]{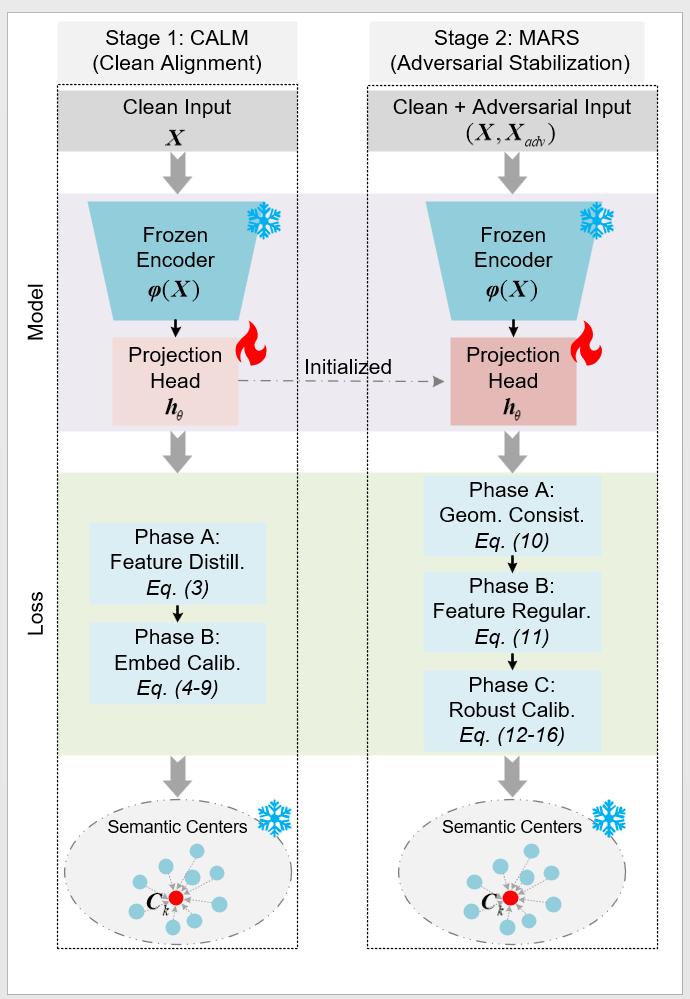}
    \caption{
    \textbf{Overview of the \name{} framework.}
    Stage~1 \textbf{CALM} performs clean feature distillation and embedding calibration with a frozen encoder $\varphi(X)$ and a lightweight projection head $h_\theta$.
    Stage~2 \textbf{MARS} refines both clean and adversarial inputs $(X, X_{\text{adv}})$ through geometric consistency, feature regularization, and robust calibration.
    The projection head $h_\theta$ has less than \textbf{1\%} of the encoder’s parameters.
}

    \label{fig:main_figure}
\end{figure}

\section{Methodology}

\noindent \textbf{Preliminaries.} %
\label{sec:bind_robustness}
Bind models~\cite{han2023imagebind,zhu2024languagebind,lyu2024unibind,lyu2024omnibind} aim to encode heterogeneous inputs into a shared embedding space. Given a modality \( m \in \mathcal{M} \), its encoder \(\phi_m: X_m \!\rightarrow\! \mathbb{R}^D\) maps an input \( x_m \) to a \( D \)-dimensional representation. A set of semantic centers
\(\{c_k\}_{k=1}^{K}\) defines the shared reference space, where each \(c_k\) is either generated from language descriptions
or jointly learned across modalities. For classification tasks, a Bind model behaves as a cosine classifier:
\begin{equation}
\setlength{\abovedisplayskip}{3pt}
\setlength{\belowdisplayskip}{3pt}
\begin{aligned}
f_k(\phi_m, x_m)
&= \cos\!\left(
\frac{\phi_m(x_m)}{\|\phi_m(x_m)\|_2},
\frac{c_k}{\|c_k\|_2}
\right), \\
\hat{y} &= \arg\max_{k} f_k(\phi_m, x_m).
\end{aligned}
\label{eq:cosine-logit}
\end{equation}
An input \(x_m^{\text{adv}}\) is adversarial if
\begin{equation}
\setlength{\abovedisplayskip}{3pt}
\setlength{\belowdisplayskip}{3pt}
\arg\max_{k} f_k(\phi_m, x_m^{\text{adv}}) \neq y,
\quad
\|x_m^{\text{adv}} - x_m\|_p \le \epsilon,
\label{eq:adv-example}
\end{equation}
where \(\epsilon\) is the perturbation radius and
\(\|\cdot\|_p\) is typically chosen as \(\ell_\infty\).
We adopt AutoAttack~\citep{croce2020reliable} as a standardized evaluation protocol under this frozen classifier setting.

As follow, we first introduce \textbf{CALM}, which focuses on improving multi-modal encoders' performance on clean samples through feature distillation and embedding calibration. We then present \textbf{MARS}, a complementary adversarial training framework designed to enhance robustness against perturbations. Together, CALM and MARS form the \textbf{\name}~framework: CALM preserves semantic alignment under clean conditions, while MARS fortifies the same encoders against adversarial attacks, including AutoAttack~\cite{} on classification and CrossFire~\cite{dou2024crossfire} attacks on retrieval tasks. 
Lastly, we introduce \textbf{CrossMaxim}, a non-targeted attack that repels representations from labels without relying on external hard-negatives.

\subsection{CALM}
\label{sec:calm}
CALM aims to have better multi-modal encoders on clean performance.
We first freeze the multi-modal encoder as teacher $\phi_m$ and learn a projection head as student $h_\theta:\mathbb{R}^D\!\rightarrow\!\mathbb{R}^D$ on top of teacher features for each modality. Given an input $x$ and its class label $y\!\in\!\{1,\dots,K\}$, the normalized teacher embedding from $\phi_m$ is
\(
t(x)=\mathrm{norm}\!\big(\phi_m(x)\big)\in\mathbb{R}^D
\),
and the student embedding is
\(
z(x)=\mathrm{norm}\!\big(h_\theta(t(x))\big)\in\mathbb{R}^D.
\)
We maintain a fixed class-center bank \(C=[c_1,\dots,c_K]\in\mathbb{R}^{D\times K}\) with column-wise normalization (\(\|c_k\|_2\!=\!1\)), constructed from language descriptions or precomputed centers
(See \S\ref{sec:bind_robustness}).

\noindent \textbf{Phase-A: Feature Distillation (FD).}
To achieve the ultimate goal, we first do feature-level knowledge distillation. Specifically, we warm up the projection head $h_\theta$ by geometrically inheriting the teacher manifold:
\begin{equation}
\setlength{\abovedisplayskip}{3pt}
\setlength{\belowdisplayskip}{3pt}
\label{eq:fd}
\mathcal{L}_{\mathrm{FD}}
=\big\|\,h_\theta\big(t(x)\big)-t(x)\,\big\|_2^2.
\end{equation}
This phase stabilizes optimization and preserves foundation knowledge within the teacher multi-modal encoders .

\noindent \textbf{Phase-B: Embedding-Guided Calibration.}
After the feature-level warm-up, the projection head $h_\theta$
is further calibrated to learn discriminative yet geometrically consistent
embeddings through an embedding-guided objective.
This phase jointly enforces
(1) Embedding-based Cross-Entropy (ECE),
(2) Angular Regularization Margin (ARM), and
(3) Counterfactual Direction Separation (CDS).
Let the normalized student embedding be
$z=\mathrm{norm}(h_\theta(t(x)))$
and the normalized class embeddings be
$C=[c_1,\dots,c_K]\!\in\!\mathbb{R}^{D\times K}$ with $\|c_k\|_2\!=\!1$.
The logit vector is
\(\ell(z)=z^{\top}C\in\mathbb{R}^K.\)

\noindent\textbf{(i) Embedding-based Cross-Entropy (ECE).}
The first objective ensures that each projected embedding
is classified according to its corresponding class embedding.
We apply a temperature-scaled cross-entropy loss:
\begin{equation}
\setlength{\abovedisplayskip}{3pt}
\setlength{\belowdisplayskip}{3pt}
\label{eq:ece}
\mathcal{L}_{\mathrm{ECE}}
=\mathrm{CE}\!\left(\tfrac{1}{\tau}\,\ell(z),\,y\right),
\end{equation}
where $\tau$ controls the softness of class probabilities.
This term preserves the teacher’s decision boundaries
while allowing mild geometric adaptation.

\noindent\textbf{(ii) Angular Regularization Margin (ARM).}
To enlarge the inter-class angular gap and
reduce vulnerability near decision boundaries,
we adopt a CosFace-style margin on the target logit
and scale all logits by a fixed factor $s$:
\begin{equation}
\setlength{\abovedisplayskip}{3pt}
\setlength{\belowdisplayskip}{3pt}
\label{eq:arm}
\tilde\ell_k =
\begin{cases}
s\!\cdot\!(\ell_y-m), & k=y,\\[2pt]
s\!\cdot\!\ell_k, & k\neq y,
\end{cases}
\qquad
\mathcal{L}_{\mathrm{ARM}}
=\mathrm{CE}(\tilde\ell,\,y).
\end{equation}
Here $m$ is the angular margin.
This constraint encourages the embedding $z$ to maintain a stable
cosine distance from its class embedding $c_y$,
effectively expanding the decision region and improving
tolerance to perturbations.

\noindent\textbf{(iii) Counterfactual Direction Separation (CDS).}
Even with angular margins, embeddings of hard negatives
can collapse under adversarial noise.
To explicitly separate them, we introduce a counterfactual step
toward the hardest negative class embedding.
Let $j=\arg\max_{k\neq y}\ell_k$ denote the most confusing class
(with gradient detached).
We perturb $z$ along the direction between class embeddings:
\begin{equation}
\setlength{\abovedisplayskip}{3pt}
\setlength{\belowdisplayskip}{3pt}
    z^{\mathrm{cf}}
=\mathrm{norm}\!\big(z+\delta\,(c_j-c_y)\big),
\end{equation}
and enforce a hinge margin $\kappa$ between its similarity to $c_y$
and to $c_j$:
\begin{equation}
\setlength{\abovedisplayskip}{3pt}
\setlength{\belowdisplayskip}{3pt}
\label{eq:cds}
\mathcal{L}_{\mathrm{CDS}}
=\big[\;\langle z^{\mathrm{cf}},c_y\rangle
-\langle z^{\mathrm{cf}},c_j\rangle+\kappa\;\big]_+.
\end{equation}
This counterfactual operation enlarges the local neighborhood
around each class embedding, mitigating overfitting to clean manifolds
and improving robustness against directional perturbations.

The overall embedding-guided calibration loss in CALM combines the three components:
\begin{equation}
\setlength{\abovedisplayskip}{3pt}
\setlength{\belowdisplayskip}{3pt}
\label{eq:total_b}
\mathcal{L}_{\mathrm{B}}
=\lambda_{\mathrm{ECE}}\mathcal{L}_{\mathrm{ECE}}
+\lambda_{\mathrm{ARM}}\mathcal{L}_{\mathrm{ARM}}
+\lambda_{\mathrm{CDS}}\mathcal{L}_{\mathrm{CDS}}.
\end{equation}
Training follows a two-phase curriculum, where a fraction
$\rho\!\in(0,1)$ of total epochs $T$ is devoted to feature distillation:
\begin{equation}
\setlength{\abovedisplayskip}{3pt}
\setlength{\belowdisplayskip}{3pt}
\label{eq:schedule}
\mathcal{L}_{\text{CALM-KD}}=
\begin{cases}
\mathcal{L}_{\mathrm{FD}}, & e \le \rho T,\\[2pt]
\mathcal{L}_{\mathrm{B}},  & e > \rho T,
\end{cases}
\qquad \rho\!\approx\!0.5.
\end{equation}


\subsection{MARS}
\label{sec:mars}

To complement the clean-oriented calibration of \textbf{CALM},
we introduce \textbf{MARS} (Multi-phase Adversarial Representation Stabilization), a three-phase adversarial training framework that enhances multi-modal encoders' robustness under perturbations while preserving semantic alignment. MARS progressively refines the projection head through:
\ding{172} Phase~A: geometric consistency,
\ding{173} Phase~B: robust feature regularization, and
\ding{174} Phase~C: robustness-aware metric calibration.
This curriculum stabilizes optimization and yields embeddings
that remain discriminative and semantically coherent
in adversarial conditions.

\noindent \textbf{Phase~A: Geometric Consistency.}
The first phase constrains the projection head to maintain
the geometric topology between clean and perturbed embeddings.
Given a clean--adversarial pair $(x, x^{\mathrm{adv}})$,
we enforce local consistency by minimizing the $L_2$ distance:
\begin{equation}
\setlength{\abovedisplayskip}{3pt}
\setlength{\belowdisplayskip}{3pt}
\label{eq:mars_phaseA}
\mathcal{L}_{\mathrm{A}}
=\|\,h_\theta(\phi_m(x)) - h_\theta(\phi_m(x^{\mathrm{adv}}))\,\|_2^2.
\end{equation}
This geometric constraint preserves the latent structure
of the embedding space and provides a stable initialization
for subsequent adversarial optimization.

\noindent \textbf{Phase~B: Robust Feature Regularization.}
Phase~B introduces distributional alignment between clean and adversarial predictions
to achieve a balanced robustness–accuracy trade-off.
Let $p_{\mathrm{adv}}$ and $p_{\mathrm{clean}}$
be the softmax probabilities of the projection head on
$x^{\mathrm{adv}}$ and $x$, respectively.
We define the regularization loss as
\begin{equation}
\setlength{\abovedisplayskip}{3pt}
\setlength{\belowdisplayskip}{3pt}
\label{eq:mars_phaseB}
\mathcal{L}_{\mathrm{B}}
=\mathrm{KL}\!\big(
p_{\mathrm{adv}}\,\|\,p_{\mathrm{clean}}^{\mathrm{det}}
\big)
+\lambda_{\mathrm{clean}}\,
\mathrm{CE}(p_{\mathrm{clean}}, y),
\end{equation}
the first term aligns adversarial predictions with detached clean distributions, and the second maintains clean supervision. This phase mitigates overfitting to noise, fostering robust and semantically consistent representations.

\noindent \textbf{Phase~C: Robustness-aware Embedding-Guided Calibration.}
The final phase reinforces discriminability through
(1) \textit{Robustness-aware Embedding-based Cross-Entropy (\textbf{R-ECE})},
(2) \textit{Robustness-aware Angular Regularization Margin (\textbf{R-ARM})}, and
(3) \textit{Robustness-aware Counterfactual Direction Separation (\textbf{R-CDS})}.
This design mirrors the structure of CALM's Phase~B
but redefines the objectives under adversarial supervision,
transforming clean calibration into robustness-oriented calibration. Let the normalized adversarial embedding be
$z=\mathrm{norm}(h_\theta(\phi_m(x^{\mathrm{adv}})))$
and the normalized class embeddings be
$C=[c_1,\dots,c_K]\!\in\!\mathbb{R}^{D\times K}$ with $\|c_k\|_2=1$.
The logits are $\ell(z)=z^\top C$.

\noindent\textbf{(i) R-ECE.}
We first align adversarial embeddings with their semantic classes
using a temperature-scaled cross-entropy:
\begin{equation}
\setlength{\abovedisplayskip}{3pt}
\setlength{\belowdisplayskip}{3pt}
\label{eq:mars_recep}
\mathcal{L}_{\mathrm{R\text{-}ECE}}
=\mathrm{CE}\!\left(\tfrac{1}{\tau}\,\ell(z),\,y\right),
\end{equation}
where $\tau$ adjusts prediction sharpness.
Unlike CALM, which focuses on clean alignment,
R-ECE optimizes classification consistency directly on adversarial embeddings.

\noindent\textbf{(ii) R-ARM.}
To expand inter-class separation and stabilize boundaries,
we modify the target logit with an angular margin $m$
and scale $s$:
\begin{equation}
\setlength{\abovedisplayskip}{3pt}
\setlength{\belowdisplayskip}{3pt}
\label{eq:mars_arm}
\tilde{\ell}_k =
\begin{cases}
s(\ell_y - m), & k=y,\\[2pt]
s\,\ell_k, & k\neq y,
\end{cases}
\qquad
\mathcal{L}_{\mathrm{R-ARM}}
=\mathrm{CE}(\tilde{\ell}, y).
\end{equation}
R-ARM encourages robust angular separation, making embeddings less sensitive to boundary perturbations.

\noindent\textbf{(iii) R-CDS.}
Finally, we enhance counterfactual robustness by repelling
the hardest negative direction.
Let $j=\arg\max_{k\neq y}\ell_k$ be the hardest negative
(with gradients detached),
and define
$z^{\mathrm{cf}}=\mathrm{norm}\!\big(z+\delta(c_j-c_y)\big)$.
We enforce a hinge margin $\kappa$:
\begin{equation}
\setlength{\abovedisplayskip}{3pt}
\setlength{\belowdisplayskip}{3pt}
\label{eq:mars_cds}
\mathcal{L}_{\mathrm{R-CDS}}
=\big[\,
\langle z^{\mathrm{cf}},c_y\rangle
-\langle z^{\mathrm{cf}},c_j\rangle
+\kappa
\,\big]_+.
\end{equation}
This counterfactual constraint expands local neighborhoods
around each class embedding, producing adversarially robust alignment.
MARS is optimized in a curriculum manner rather than summing all phases at once.
Let $T$ be the total epochs and $(\rho_A,\rho_B,\rho_C)$ be the phase ratios
with $\rho_A{+}\rho_B{+}\rho_C{=}1$.
Define the boundaries $T_A{=}\rho_A T$ and $T_B{=}( \rho_A{+}\rho_B )T$.
The phase-wise objective at epoch $e\in\{1,\dots,T\}$ is
\begin{equation}
\setlength{\abovedisplayskip}{3pt}
\setlength{\belowdisplayskip}{3pt}
\label{eq:mars_schedule}
\mathcal{L}_{\text{MARS}}(e)=
\begin{cases}
\mathcal{L}_{\mathrm{A}}, & e \le T_A,\\[3pt]
\mathcal{L}_{\mathrm{B}}, & T_A < e \le T_B,\\[3pt]
\mathcal{L}_{\mathrm{C}}, & e > T_B,
\end{cases}
\end{equation}
where Phase~C combines robustness-aware metric terms:
\begin{equation}
\setlength{\abovedisplayskip}{3pt}
\setlength{\belowdisplayskip}{3pt}
\label{eq:mars_phaseC_combo}
\mathcal{L}_{\mathrm{C}}
=\alpha_{\mathrm{R\text{-}ECE}}\,\mathcal{L}_{\mathrm{R\text{-}ECE}}
+\alpha_{\mathrm{R\text{-}ARM}}\,\mathcal{L}_{\mathrm{R\text{-}ARM}}
+\alpha_{\mathrm{R\text{-}CDS}}\,\mathcal{L}_{\mathrm{R\text{-}CDS}}.
\end{equation}


\subsection{CrossMaxim Attack}
\label{sec:crossmaxim}
Most targeted attacks push samples toward specific negatives, like a sniper aiming at one mismatch. This sharply drops Top-1 accuracy but leaves many correct pairs in Top-5.
We propose \textbf{CrossMaxim}, a non-targeted “machine-gun” attack that broadly repels embeddings from their true matches, degrading overall alignment without using hard-negative labels.

\noindent \textbf{Objective.}
Given a clean sample \(x\) from modality \(a\)
and its paired embedding \(t=\phi_b(x_b)\) from modality \(b\),
CrossMaxim generates an adversarial example \(x^{\mathrm{adv}}\)
that maximizes the cosine distance between
their embeddings:
\begin{equation}
\setlength{\abovedisplayskip}{3pt}
\setlength{\belowdisplayskip}{3pt}
\label{eq:crossmaxim_obj}
x^{\mathrm{adv}}
=\arg\max_{\|x'-x\|_\infty\le\epsilon}
\big[-\cos(\phi_a(x'),\,t)\big].
\end{equation}
Equivalently, the inner optimization minimizes
the cosine similarity
\(
\mathcal{L}_{\mathrm{CM}}=-\cos(\phi_a(x^{\mathrm{adv}}),t)
\),
driving \(x^{\mathrm{adv}}\) away from its true caption/audio partner.
This objective differs from targeted CrossFire, which instead
\emph{increases} similarity with a chosen hard negative.

\noindent \textbf{Optimization.}
We adopt a normalized PGD routine with step size \(\alpha=\epsilon/T\) over \(T\) iterations:
\begin{equation}
\setlength{\abovedisplayskip}{3pt}
\setlength{\belowdisplayskip}{3pt}
\label{eq:crossmaxim_pgd}
x^{(t+1)}
=\Pi_{[0,1]\!\cap\!\mathcal{B}_\infty(x,\epsilon)}
\!\big(x^{(t)}+\alpha\,\mathrm{sign}(\nabla_x \mathcal{L}_{\mathrm{CM}})\big),
\end{equation}
where the projection \(\Pi\) enforces valid pixel ranges and
the $\ell_\infty$ constraint.
Gradients are computed through the white-box encoder of the
attacked modality (image or audio branch).
To ensure reproducibility, all random seeds and
\texttt{cuDNN} backends are fixed to deterministic mode.

\noindent \textbf{Implementation details.}
CrossMaxim is implemented as a lightweight plug-in built upon
the same retrieval interface used by AutoAttack.
We use per-channel normalization with mean
\((0.481,0.458,0.408)\) and std
\((0.269,0.261,0.276)\) for vision inputs,
and a single-channel variant for audio.
For each iteration we
(1) forward the current $x^{(t)}$ to obtain embeddings,
(2) compute cosine loss $-\cos(z_{\mathrm{adv}},t)$,
(3) update $x^{(t)}$ by gradient ascent,
and (4) project back to the valid $\ell_\infty$ ball.
Typical settings use $\epsilon\!\in\!\{2,4,8\}/255$ and
$T\!=\!20$ iterations.

\noindent \textbf{Validation and diagnostics.}
After generation, we verify that the perturbation magnitude
satisfies the normalized bound
$\|x^{\mathrm{adv}}-x\|_\infty\!\le\!\epsilon$
and that the cosine similarity to the clean embedding
has decreased:
\(
\Delta\!\cos
=\cos(\phi_a(x^{\mathrm{adv}}),t)
-\cos(\phi_a(x),t) < 0.
\)
A successful attack thus corresponds to a negative
$\Delta\!\cos$, indicating that the adversarial representation
has been effectively repelled from the correct counterpart.

\noindent \textbf{Discussion.}
Compared with CrossFire, CrossMaxim
(1) requires no hard-negative metadata,
(2) generalizes naturally to all modality pairs
(e.g., image–text, audio–text, event–text),
and (3) yields substantially higher Attack Success Rates (ASR)
under the same $\epsilon$ and iteration budget.
Empirically, it reduces retrieval R@1 by an additional
$8$–$12$ points over CrossFire on average,
revealing previously overlooked vulnerability modes
in Bind-style encoders.

\section{Experiments}
\label{sec:exp}
\subsection{Zero-Shot Classification Robustness}

\textbf{Datasets.}
We evaluate our method across six modalities using standard benchmarks:
\textbf{(1) Image}: ImageNet-1K~\cite{deng2009imagenet} and Places365~\cite{lopez2020semantic};
\textbf{(2) Video}: MSR-VTT~\cite{xu2016msr} and UCF-101~\cite{soomro2012ucf101};
\textbf{(3) Event}: N-ImageNet1K~\cite{dou2023adversarial};
\textbf{(4) Point Cloud}: ModelNet40~\cite{wu20153d};
\textbf{(5) Thermal}: LLVIP~\cite{jia2021llvip};
and \textbf{(6) Audio}: ESC-50~\cite{piczak2015esc}.
We subsample 2000 evaluation examples for N-ImageNet1K and 1000 examples for all other datasets to ensure at least two samples per class.
Training sets are five times the evaluation size, except for ESC-50, where the full dataset is used due to its smaller size.

\noindent \textbf{\ding{172} Adversarial Attack for classification.}
We use AutoAttack~\cite{croce2020reliable} as our standard evaluation, running its four default attacks (\texttt{apgd-ce}, \texttt{apgd-dlr}, \texttt{fab-t}, \texttt{square}) with $\ell_\infty$ budgets $\epsilon\in{2,4,8}/255$. 
For LLVIP (binary), targeted attacks (\texttt{apgd-dlr}, \texttt{fab-t}) are excluded; only \texttt{apgd-ce} and \texttt{square} are used. During adversarial training we use \texttt{apgd-ce} with $\epsilon=8/255$ for efficiency and generalization, following evidence that high-$\epsilon$ single-attack training transfers across budgets~\cite{rebuffi2021fixing, gowal2020uncovering, zhang2019theoretically}. (See appendix)
We evaluate classification adversarial robustness across three Bind models: \textit{ImageBind}~\cite{girdhar2023imagebind}, \textit{LanguageBind}~\cite{zhu2023languagebind}, and \textit{UniBind}~\cite{lyu2024unibind}. 
Table in appendix summarizes clean and adversarial accuracies under AutoAttack ($\epsilon \in {2, 4, 8}/255$). All Bind models show strong accuracy on clean data but collapse under perturbations. Robustness differs across modalities, showing that unified representations alone don’t guarantee cross-modal resilience.

\begin{figure}[t]
\centering
\includegraphics[width=\linewidth]{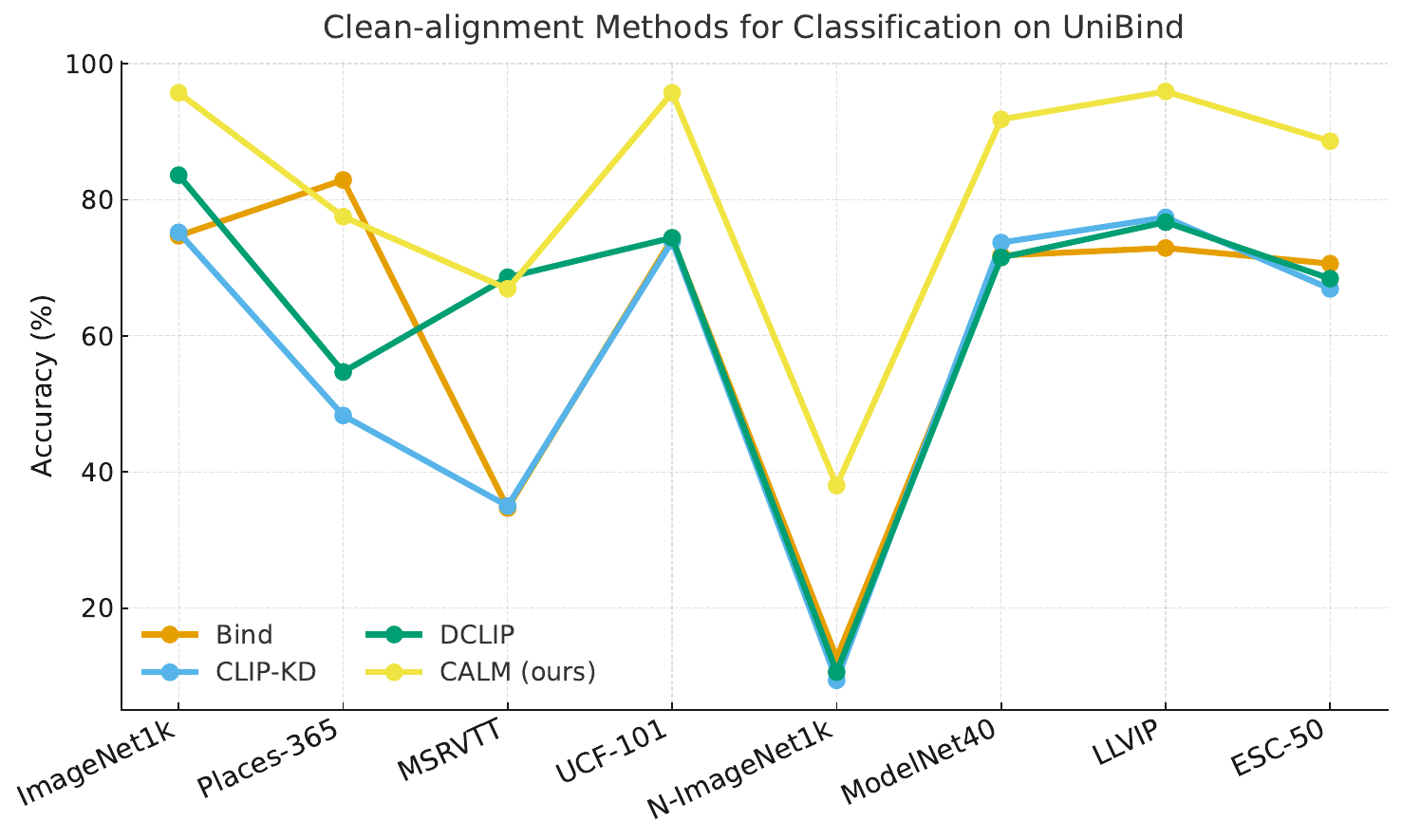}
\caption{Comparison of clean-alignment methods across eight datasets on \textit{UniBind}.
Each line represents a distillation-based baseline (Bind, CLIP-KD, DCLIP) or our CALM method.}
\vspace{-8pt}
\label{fig:main_kd_cls_line}
\vspace{-8pt}
\end{figure}

\noindent \textbf{\ding{173} Clean-oriented Alignment for classification.}
We evaluate CALM in the clean-only alignment stage and compare it with distillation-based baselines (CLIP-KD, DCLIP, Meta-Adapter).
For \textit{UniBind}, a visual summary is shown in Fig.~\ref{fig:main_kd_cls_line}; detailed per-dataset Top-1 results and the corresponding training time are reported in Appendix Tables.
Across models and datasets, CALM attains the highest or statistically comparable accuracy while being more time-efficient.

\noindent \textbf{\ding{174} Adversarial  Training for Classification.}
\label{sec:cls_at}
Table~\ref{tab:cls_main} summarizes the classification performance of \textit{CALMARS} and prior adversarial training baselines across eight datasets under perturbation budgets $\epsilon\!\in\!\{2,4,8\}/255$. 
The results reveal several clear trends.
First, \textit{CALMARS achieves the \textbf{best} overall robustness--accuracy trade-off} among all methods. 
On the large-scale visual datasets (\textbf{ImageNet1K} and \textbf{Places-365}), CALMARS matches or slightly trails the best clean accuracy while maintaining strong robustness. 
Second, in non-image modalities (N-ImageNet1K, ModelNet40, LLVIP, ESC-50, MSR-VTT), CALMARS consistently \textbf{\textit{outperforms all}} prior approaches under both clean and adversarial settings. 
Overall, CALMARS shows \textbf{the most consistent robustness across modalities}, preserving high clean accuracy and avoiding the sharp degradation seen in other methods as perturbations grow.
These results demonstrate that CALMARS’s multi-stage adversarial refinement effectively improves cross-modal stability without harming general accuracy.

\begin{table}[t]
\centering
\scriptsize
\setlength{\tabcolsep}{5pt}
\renewcommand{\arraystretch}{0.85}
\caption{Clean-alignment methods  for retrieval on \textit{ImageBind} and \textit{LanguageBind}. 
Numbers are Recall@1 / Recall@5 (\%). 
Best per VLM and dataset in \textbf{bold}.}
\label{tab:kd_rtl}
\resizebox{\linewidth}{!}{
\begin{tabular}{l l | cc | cc}
\toprule
\multirow{2}{*}{\textbf{VLM}} & \multirow{2}{*}{\textbf{Encoder}} &
\multicolumn{2}{c|}{\textbf{Image}} &
\multicolumn{2}{c}{\textbf{Audio}} \\
 &  & MSCOCO & Flickr8k & AudioCaps & Clotho \\
\midrule
\multirow{3}{*}{ImageBind}
& Bind      & 65.7/91.1   & 71.4/92.8  & 8.1/25.9   & 5.5/20.3\\
& CLIP-KD      & 63.8/90   & 68.6/92.1  & 6.9/25.9   & 6.3/20 \\
& DCLIP        & 1.6/5.9   & 1.1/3.7    & 1.3/5.7    & 1.2/4 \\
\rowcolor{gray!10}& \textbf{CALM (ours)} & \textbf{76.3}/\textbf{96.7} & \textbf{77.5}/\textbf{96.8} & \textbf{26.6}/\textbf{58.5} & \textbf{25.7}/\textbf{57.8} \\
\midrule
\multirow{4}{*}{LanguageBind}
& Bind      & 63.7/90.5   & 67.3/90.8  & 14.2/40.4   & 17.6/40 \\
& CLIP-KD      & 60.9/89.5 & 64.8/89.9  & 10.1/33.9  & 9.7/26.7 \\
& DCLIP        & 1.2/5     & 1.2/3.7    & 0.7/2.3    & 1.1/3.7 \\
& Meta Adapter & 55.2/84.2 & 56.4/84.5  & 9.8/30.3   & 9.6/26 \\
\rowcolor{gray!10}& \textbf{CALM (ours)} & \textbf{72.3}/\textbf{96.1} & \textbf{75.7}/\textbf{96.1} & \textbf{30.8}/\textbf{67.2} & \textbf{25.5}/\textbf{55.5} \\
\bottomrule
\end{tabular}}
\end{table}
\begin{table}[t]
\centering
\scriptsize
\setlength{\tabcolsep}{5pt}
\renewcommand{\arraystretch}{0.85}
\caption{Cross-modal retrieval between \textit{ImageNet1K} (image) and \textit{N-ImageNet1K} (event) at $\epsilon{=}4/255$. 
Each method uses its adversarially trained MLPs from Sec.~\ref{sec:cls_at} to evaluate 100 paired image–event samples. 
Numbers are Recall@1 / Recall@5 (\%). \textbf{bold} indicates the best-performing method in each retrieval direction.}
\vspace{-4pt}
\label{tab:xmodal}
\begin{tabular}{lcc}
\toprule
\textbf{Method} & \textbf{image$\rightarrow$event} & \textbf{event$\rightarrow$image} \\
\midrule
Bind   & 19.1/32.4& 16.2/30.9\\
FARE   & 29.9/35.3& 26.4/44.1\\
TeCoA  & 25/33.8& 23.5/36.8\\
LAAT   & 36.8/39.7& 0/3\\
TRADES & 23.5/30.8   & 25/45.6\\
\rowcolor{gray!10}\textbf{CALMARS} &              \textbf{36.8}/\textbf{36.8}&             \textbf{36.7}/\textbf{51.8}\\
\bottomrule
\end{tabular}
\vspace{-14pt}
\end{table}

\begin{table*}[t]
\centering
\scriptsize
\setlength{\tabcolsep}{2pt}
\renewcommand{\arraystretch}{0.75}
\caption{Classification Acc (\%) under clean and $\ell_\infty$ budgets (2/4/8) across eight datasets. \textbf{Bold} indicates the best performance per column.}
\label{tab:cls_main}
\resizebox{\textwidth}{!}{%
\begin{tabular}{@{}l
C{10mm}C{8mm}C{8mm}C{8mm}|
C{10mm}C{8mm}C{8mm}C{8mm}|
C{10mm}C{8mm}C{8mm}C{8mm}|
C{10mm}C{8mm}C{8mm}C{8mm}@{}}
\toprule
\multirow{2}{*}{\textbf{Method}} &
\multicolumn{4}{c|}{\textbf{ImageNet1k}} &
\multicolumn{4}{c|}{\textbf{Places-365}} &
\multicolumn{4}{c|}{\textbf{MSR-VTT}} &
\multicolumn{4}{c}{\textbf{UCF-101}} \\
\cmidrule(lr){2-5}\cmidrule(lr){6-9}\cmidrule(lr){10-13}\cmidrule(lr){14-17}
& Clean & 2/255 & 4/255 & 8/255
& Clean & 2/255 & 4/255 & 8/255
& Clean & 2/255 & 4/255 & 8/255
& Clean & 2/255 & 4/255 & 8/255 \\
\midrule
UniBind          & 74.7 & 0.0 & 0.0 & 0.0 & 49.9 & 0.0 & 0.0 & 0.0 & 34.7 & 17 & 15.5 & 13.7 & 74.3 & 20.3 & 16.3 & 12.5  \\
+FARE          & 95.6 & 34.3 & 32.9 & 32.2 & 75.2 & 38.0 & 37.9 & 37.7 & 65  & 62.6  & 62.7  & 62  & 41.4 & 21.7 & 21.6 & 19.2 \\
+TeCoA         & 91.5 & 31.6 & 30.4 & 30.0 & 73.1 & 35.8 & 35.7 & 35.7 & 61.7  & 61.5  & 62.1  & 62.2  & 52.3 & 54.0 & 56.4 & 56.6 \\
+LAAT          & \textbf{95.9} & \textbf{35.2} & \textbf{33.1} & \textbf{32.3} & \textbf{77.0} & \textbf{39.4} & \textbf{39.3} & \textbf{39.2} & 62.7 & 59.5 & 59.7 & 59.5 & 7.4 & 7.2 & 7.1 & 7.1 \\
+TRADES        & 93.5 & 32.3 & 30.8 & 30.3 & 67.9 & 32.0 & 31.9 & 31.9 & 62.5 & 58.6 & 58.8 & 59.2 & 59.5 & 51.2 & 52.6 & 53.4 \\
\rowcolor{gray!10}+\textbf{CALMARS} & 95.3 & 34.6 & 32.9 & 32.1 & 76.3 & 38.5 & 38.3 & 38.3 & \textbf{65.8} & \textbf{63} & \textbf{63} & \textbf{62.5} & \textbf{63.2} & \textbf{57.6} & \textbf{58.5} & \textbf{59.1} \\
\midrule
\multirow{2}{*}{\textbf{Method}} &
\multicolumn{4}{c|}{\textbf{N-ImageNet1K}} &
\multicolumn{4}{c|}{\textbf{ModelNet40}} &
\multicolumn{4}{c|}{\textbf{LLVIP}} &
\multicolumn{4}{c}{\textbf{ESC-50}} \\
\cmidrule(lr){2-5}\cmidrule(lr){6-9}\cmidrule(lr){10-13}\cmidrule(lr){14-17}
& Clean & 2/255 & 4/255 & 8/255
& Clean & 2/255 & 4/255 & 8/255
& Clean & 2/255 & 4/255 & 8/255
& Clean & 2/255 & 4/255 & 8/255 \\
\midrule
UniBind          & 12.7 & 6.2 & 4.8 & 2.7 & 71.8 & 0.0 & 0.0 & 0.0 & 72.9 & 8.9 & 1.6 & 0.2 & 70.6 & 7.7 & 3.2 & 1.5 \\
+FARE          & 13.4 & 11.9 & 11.5 & 10.4 & 44.1 & 36.3 & 41.6 & 43.2 & \textbf{96.4} & \textbf{93.7} & \textbf{94.7} & 94.9 & 51.0 & 28.5 & 35.7 & 38.9 \\
+TeCoA         & 17.8 & 16.4 & 16.1 & 14.9 & 31.6 & 39.5 & 43.4 & 45.0 & 78.9 & 89.5 & 93.3 & 94.9 & 15.8 & 25.1 & 37.4 & 44.9 \\
+LAAT          & 1.35 & 1.3 & 1.3 & 1.4 & 39.0 & 34.7 & 36.7 & 37.8 & 94.4 & 91.1 & 92.0 & 93.0 & 7.0 & 7.0 & 10.7 & 14.7 \\
+TRADES        & 19.3 & 17.9 & 17.0 & 16.0 & 45.9 & 39.9 & 43.8 & \textbf{45.4} & 94.3 & 90.8 & 92.0 & 92.9 & 35.5 & 16.7 & 20.0 & 21.3 \\
\rowcolor{gray!10}+\textbf{CALMARS} & \textbf{19.4} & \textbf{17.9} & \textbf{17.3} & \textbf{16.0} & \textbf{45.7} & \textbf{41.7} & \textbf{44.2} & 45.0 & 96.1 & 93.6 & 94.6 & \textbf{95.0} & \textbf{54.1} & \textbf{45.3} & \textbf{48.3} & \textbf{50.3} \\
\bottomrule
\end{tabular}%
}
\end{table*}

\subsection{Cross-Modal Retrieval Robustness }

\textbf{Datasets and Modalities.}
We evaluate \name\~on four standard cross-modal benchmarks spanning visual and audio domains.
For the \textbf{image–text} domain, we use MSCOCO~\cite{lin2014microsoft} and Flickr8k~\cite{hodosh2013framing}, both containing paired image–caption data for retrieval and captioning tasks.
For the \textbf{audio–text} domain, we use AudioCaps~\cite{kim2019audiocaps} and Clotho~\cite{drossos2020clotho}, which provide natural audio clips paired with human-written captions.
We uniformly sample 1,000 examples from each dataset for evaluation and use training sets five times larger for diversity.
All datasets are used directly, without task-specific fine-tuning, to preserve the zero-shot evaluation setting.

\noindent \textbf{\ding{172} Adversarial Attack for Retrieval.}
We evaluate retrieval robustness using two retrieval-specific white-box attacks: a targeted hard-negative alignment attack (\textbf{CrossFire}) and an untargeted similarity-minimizing attack (\textbf{CrossMaxim}). 
For adversarial \textbf{\textit{training}}, we use a single-attack configuration with $\epsilon=8/255$ (for both methods) to ensure efficiency and better transferability across budgets, while for \textbf{evaluation} we test with $\epsilon\in\{2,4,8\}/255$. 
For the detailed formulation of CrossMaxim, please refer to Sec.~\ref{sec:crossmaxim}.
We evaluate cross-modal retrieval robustness across: \textit{ImageBind}~\cite{girdhar2023imagebind} and\textit{LanguageBind}~\cite{zhu2023languagebind}. 
UniBind~\cite{lyu2024unibind} is excluded because it relies on predefined modality centers to maintain a unified embedding space and does not natively support the retrieval datasets used in this study. 

Table in appendix reports cross-modal retrieval robustness under perturbation budgets $\epsilon\!\in\!\{2,4,8\}/255$.
As $\epsilon$ increases, both \textit{ImageBind} and \textit{LanguageBind} show sharp accuracy drops across datasets, confirming that even small perturbations severely disrupt cross-modal alignment.
\textit{CrossMaxim} causes much stronger degradation than \textit{CrossFire}, particularly in high-rank metrics.
While CrossFire moderately reduces top-1 accuracy, it leaves top-5 retrieval largely intact—for example, R\@5 stays above \$60
In contrast, CrossMaxim nearly collapses retrieval entirely, driving R\@5 below \$1
This stark gap shows that \textit{CrossMaxim} not only breaks top-1 matches but also removes all plausible top-5 candidates, revealing deeper vulnerabilities in the embedding space.
Overall, these results demonstrate that \textit{CrossMaxim causes catastrophic retrieval failure}, far exceeding CrossFire in both intensity and consistency, and thus serves as a stronger and more diagnostic robustness benchmark across modalities.

\newcommand{\g}[1]{\cellcolor{gray!10}{#1}}

\begin{table*}[h!]
\centering
\scriptsize
\setlength{\tabcolsep}{1pt}
\renewcommand{\arraystretch}{0.95}
\caption{Retrieval performance (Recall@1 / Recall@5, \%) under clean and $\ell_\infty$-bounded settings. Best Recall@1 is highlighted in \textbf{bold}.}
\label{tab:bind_retrieval}
\resizebox{\textwidth}{!}{
\begin{tabular}{l l l| cccc|cccc|cccc|cccc}
\toprule
\multirow{2}{*}{\textbf{Eval}} & \multirow{2}{*}{\textbf{VLM}} & \multirow{2}{*}{\textbf{Encoder}} &
\multicolumn{4}{c|}{\textbf{MSCOCO}} &
\multicolumn{4}{c|}{\textbf{Flickr8k}} &
\multicolumn{4}{c|}{\textbf{AudioCaps}} &
\multicolumn{4}{c}{\textbf{Clotho}} \\
\cmidrule(lr){4-7}\cmidrule(lr){8-11}\cmidrule(lr){12-15}\cmidrule(lr){16-19}
 &  &  & \textit{Clean} & \multicolumn{3}{c|}{$\ell_\infty$}
 & \textit{Clean} & \multicolumn{3}{c|}{$\ell_\infty$}
 & \textit{Clean} & \multicolumn{3}{c|}{$\ell_\infty$}
 & \textit{Clean} & \multicolumn{3}{c}{$\ell_\infty$} \\
 &  &  &
 & \raisebox{0.3ex}{\scriptsize 2}/\raisebox{-0.3ex}{\scriptsize 255}
 & \raisebox{0.3ex}{\scriptsize 4}/\raisebox{-0.3ex}{\scriptsize 255}
 & \raisebox{0.3ex}{\scriptsize 8}/\raisebox{-0.3ex}{\scriptsize 255}
 &  & \raisebox{0.3ex}{\scriptsize 2}/\raisebox{-0.3ex}{\scriptsize 255}
 & \raisebox{0.3ex}{\scriptsize 4}/\raisebox{-0.3ex}{\scriptsize 255}
 & \raisebox{0.3ex}{\scriptsize 8}/\raisebox{-0.3ex}{\scriptsize 255}
 &  & \raisebox{0.3ex}{\scriptsize 2}/\raisebox{-0.3ex}{\scriptsize 255}
 & \raisebox{0.3ex}{\scriptsize 4}/\raisebox{-0.3ex}{\scriptsize 255}
 & \raisebox{0.3ex}{\scriptsize 8}/\raisebox{-0.3ex}{\scriptsize 255}
 &  & \raisebox{0.3ex}{\scriptsize 2}/\raisebox{-0.3ex}{\scriptsize 255}
 & \raisebox{0.3ex}{\scriptsize 4}/\raisebox{-0.3ex}{\scriptsize 255}
 & \raisebox{0.3ex}{\scriptsize 8}/\raisebox{-0.3ex}{\scriptsize 255} \\
\midrule
\multirow{10}{*}{\rotatebox[origin=c]{90}{\textbf{CrossFire}}}
& \multirow{5}{*}{\rotatebox[origin=c]{90}{\textbf{ImageBind}}}
& Bind   & 65.7/91.1 & 8.7/80.6 & 5.8/74.4 & 5.3/67.4
& 71.4/92.8 & 5.1/79.9 & 3.0/72.6 & 2.8/65.0
& 8.1/25.9 & 0.7/20.3 & 0.5/18.9 & 0.3/18.7
& 5.5/20.3 & 0.4/12.9 & 0.3/12.8 & 0.3/11.6 \\
& & +FARE   & 63.9/91.2 & 5.5/79.0 & 1.2/72.7 & 0.3/67.4
& 69.9/93.4 & 5.2/80.0 & 0.7/71.8 & 0.2/64.3
& 6.8/23.6 & 3.5/14.2 & 3.2/14.0 & 3.4/13.8
& 6.3/19.0 & 3.6/12.2 & 2.8/12.4 & 2.9/11.2 \\
& & +TeCoA  & 37.4/74.3 & 19.7/69.2 & 12.7/64.9 & 8.2/59.6
& 41.9/73.1 & 24.9/73.1 & 17.2/63.1 & 12.7/58.7
& 6.2/21.5 & 4.8/18.1 & 5.1/17.7 & 4.3/17.6
& 3.4/15.0 & 2.4/11.2 & 2.1/10.4 & 2.5/11.7 \\
& & +LAAT   & 31.4/63.5 & 15.4/53.4 & 11.1/49.1 & 8.4/46.6
& 35.7/68.7 & 16.9/58.3 & 9.5/51.5 & 5.5/47.8
& 3.8/15.1 & 2.3/10.6 & 2.2/10.8 & 2.6/11.0
& 1.8/8.4 & 2.6/6.6 & 1.7/6.3 & 1.7/6.6 \\
& & +TRADES & 51.9/84.1 & 21.7/73.5 & 11.9/68.0 & 7.6/63.7
& 58.4/89.9 & 23.5/75.6 & 12.5/71.4 & 7.6/63.7
& 10.8/34.2 & 4.6/19.4 & 4.7/18.4 & 3.7/17.3
& 6.8/19.0 & 3.3/14.6 & 3.3/14.6 & 3.4/13.5 \\
& & \g{\textbf{CALMARS}} & \g{\textbf{83.8}/\textbf{98.5}} & \g{\textbf{32.9}/\textbf{86.7}} & \g{\textbf{19.9}/\textbf{83.0}} & \g{\textbf{11.7}/\textbf{77.2}}
& \g{\textbf{81.8}/\textbf{97.6}} & \g{\textbf{36.8}/\textbf{88.7}} & \g{\textbf{21.6}/\textbf{82.1}} & \g{\textbf{11.3}/\textbf{75.9}}
& \g{\textbf{31.5}/\textbf{64.6}} & \g{\textbf{10.3}/\textbf{31.1}} & \g{\textbf{9.9}/\textbf{29.8}} & \g{\textbf{9.1}/\textbf{29.3}}
& \g{\textbf{38.9}/\textbf{68.9}} & \g{\textbf{11.1}/\textbf{28.3}} & \g{\textbf{9.7}/\textbf{27.7}} & \g{\textbf{9.9}/\textbf{25.8}} \\
\cmidrule(lr){2-19}
& \multirow{5}{*}{\rotatebox[origin=c]{90}{\textbf{LanguageBind}}}
& Bind   & 63.7/90.5 & 1.1/74.1 & 0.2/68   & 0.1/63.4 & 67.3/90.8 & 0.3/71.9 & 0/67.3   & 0/60.1   & 4.2/40.4 & 0.5/28   & 0.3/24.6 & 0.2/24.3 & 17.6/40  & 0.3/27.4 & 0.1/25.9 & 0.1/24.4 \\
& & FARE   & 59.5/89.9 & 1.9/73.7 & 0.2/66.8 & 0/63.3 & 64.1/90.3 & 0.7/70.3 & 0.4/62.3 & 0.3/55.5 & 14.3/39.8 & 1.4/25.3 & 0.4/22.9 & 0.2/22.9 & 16.3/39.3 & 0.4/26.3 & 0.3/24.1 & 0.1/22.5 \\
& & +TeCoA  & 32.7/67.8 & 16.1/61.2 & 11.6/58 & 7.0/55.4 & 36.2/68.0 & 14.2/59 & 9.0/57.6 & 7.7/51.7 & 10.2/31.9 & 8.6/32.7 & 7.0/30.5 & 5.3/29.6 & 8.3/29.3 & 3.4/29.0 & 3.1/26.9 & 2.3/22.1 \\
& & +LAAT   & 26.4/58.2 & 12.1/47.6 & 10.6/45.7 & 8.5/42.5 & 28.6/59.6 & 10.6/45.6 & 7.7/41.7 & 6.9/36.9 & 12.2/37.5 & 5.7/27.4 & 4.5/24.9 & 3.8/23.9 & 5.4/22.3 & 3.5/16.3 & 3.1/14.5 & 2.9/14.9 \\
& & +TRADES & 47.8/83.2 & 15.5/67.1 & 9.7/54.9  & 7.4/49.7  & 54.3/86.1 & 11.6/68.5 & 6.2/64.1 & 5.7/63.4 & 22.8/54.9 & 9.3/37.6 & 8.1/32.4 & 7.9/36.5 & 15.4/41.2 & 3.5/26.3 & 3.3/23.6 & 2.9/23.3 \\
& & \g{\textbf{CALMARS}} & \g{\textbf{74.9}/\textbf{95.7}} & \g{\textbf{22.5}/\textbf{83.8}} & \g{\textbf{14.2}/\textbf{77.2}} & \g{\textbf{7.1}/\textbf{72.9}}
& \g{\textbf{75.5}/\textbf{95.5}} & \g{\textbf{19.8}/\textbf{82.1}} & \g{\textbf{10.9}/\textbf{74.6}} & \g{\textbf{7.5}/\textbf{69.3}}
& \g{\textbf{29.9}/\textbf{62.6}} & \g{\textbf{12.8}/\textbf{41.1}} & \g{\textbf{9.2}/\textbf{37.6}} & \g{\textbf{7.9}/\textbf{36.5}}
& \g{\textbf{37.4}/\textbf{68.4}} & \g{\textbf{6.7}/\textbf{19.6}} & \g{\textbf{5.8}/\textbf{17.4}} & \g{\textbf{4.4}/\textbf{14.1}} \\
\midrule
\multirow{10}{*}{\rotatebox[origin=c]{90}{\textbf{CrossMaxim}}}
& \multirow{5}{*}{\rotatebox[origin=c]{90}{\textbf{ImageBind}}}
& Bind   & 65.7/91.1 & 3/12.3   & 0.6/2.1  & 0/0.1   & 71.4/92.8 & 4.4/13.3 & 0.8/2.9  & 0.1/0.2 & 8.1/25.9 & 0/0     & 0/0     & 0/0     & 5.5/20.3 & 0/0     & 0/0     & 0/0 \\
& & +FARE   & 63.6/90.9 & 4.9/18.4 & 1.2/5.2 & 0.7/2.1 & 69.4/92.5 & 4.9/18.4 & 1.2/5.2 & 0.7/2.1 & 7/24.3   & 1.6/4.9 & 1.2/3.8 & 0.3/2.4 & 6.2/18.9 & 1.7/4.7 & 1/3.7 & 0/0 \\
& & +TeCoA  & 41.9/76.9 & 7.3/25.2 & 2.8/10.9 & 1.5/4.1 & 46.5/79.5 & 7.6/22.3 & 2.6/9.6 & 0.6/2.6 & 4.6/17.7 & 2.2/9.4 & 2.3/8.3 & 2.1/7.8 & 3.2/12.9 & 2/8.5   & 1.3/7.5 & 0.8/5.3 \\
& & +LAAT   & 22.3/52.9 & 6.5/21   & 3.2/11.1 & 1/5.1   & 22/51.3   & 4.2/17   & 1.9/8.6  & 0.5/3   & 2/10.4   & 1/4.3   & 0.7/3   & 0.8/3.7 & 0.8/6.8  & 0.7/4.3 & 0.4/4.7 & 0.2/3.8 \\
& & +TRADES & 54.2/84.9 & 9.1/27.4 & 2.8/11.3 & 1/3.6   & 62.1/90.2 & 9.5/24.5 & 2.7/9.1  & 0.2/1.6 & 9.7/35.8 & 2.3/10  & 2.3/8.9 & 2.5/7.3 & 7.6/21.7 & 2.3/7.6 & 1.9/5.9 & 1.6/5 \\
& & \g{\textbf{CALMARS}} & \g{\textbf{81.2}/\textbf{97.4}} & \g{\textbf{20.5}/\textbf{44.2}} & \g{\textbf{6.9}/\textbf{19}} & \g{\textbf{2.4}/\textbf{6.1}}
& \g{\textbf{82.8}/\textbf{97.9}} & \g{\textbf{17.9}/\textbf{38.4}} & \g{\textbf{8.0}/\textbf{23.5}} & \g{\textbf{6.1}/\textbf{12.7}}
& \g{\textbf{32.1}/\textbf{64.1}} & \g{\textbf{5.5}/\textbf{18.9}} & \g{\textbf{4.1}/\textbf{15.4}} & \g{\textbf{3.5}/\textbf{12.7}}
& \g{\textbf{37.4}/\textbf{68.4}} & \g{\textbf{6.7}/\textbf{19.6}} & \g{\textbf{5.8}/\textbf{17.4}} & \g{\textbf{4.4}/\textbf{14.1}} \\
\cmidrule(lr){2-19}
& \multirow{5}{*}{\rotatebox[origin=c]{90}{\textbf{LanguageBind}}}
& Bind   & 63.7/90.5 & 1.8/6.3  & 0.6/1.5 & 0.4/0.5 & 67.3/90.8 & 3.7/19.8 & 0/6.7 & 0/6.1 & 4.2/40.4 & 0/0   & 0/0   & 0/0   & 17.6/40  & 0/0   & 0/0   & 0/0 \\
& & FARE   & 59.5/89.5 & 3.2/11.2 & 1.7/4.2 & 0.9/2.2 & 64.5/90.2 & 2.3/8.4  & 0.5/2.9 & 0.2/0.8 & 13.6/38.9 & 0/0.1 & 0/0.2 & 0/0.6 & 16/40.5 & 0/0.1 & 0/0.5 & 0/0 \\
& & TeCoA  & 37.2/72.5 & 5.1/16.2 & 2.1/7.1 & 0.9/3.1 & 36.6/71.3 & 4.2/13.7 & 1.7/6.1 & 0.5/2 & 2.5/10.9 & \textbf{0.8}/\textbf{3.6} & \textbf{0.9}/\textbf{3.3} & \textbf{0.6}/\textbf{3.2} & 3.4/10.3 & \textbf{2}/\textbf{1.9}  & \textbf{2}/\textbf{1.2}  & \textbf{2}/\textbf{1.1} \\
& & LAAT   & 18.1/44.6 & 4.6/15.6 & 1.7/7.6 & 0.9/4.6 & 16.6/42.9 & 2.5/10   & 1/4.8  & 0.7/1.7 & 0.6/1.8  & 0.2/0.8 & 0/0.8  & 0/0 & 0.9/5.4  & 0.1/0.7 & 0/0.6 & 0/0 \\
& & TRADES & 51.6/84   & 4.9/18.5 & 2.4/7.3 & 1/4.8   & 56.2/88.1 & 5.2/14.6 & 1.8/4.8 & 1/1.7   & 23.2/56.1 & 0/0   & 0/0   & 0/0   & 16.4/42.5 & 0/0.5 & 0/0.1 & 0/0 \\
& & \g{\textbf{CALMARS}} & \g{\textbf{79.3}/\textbf{98.2}} & \g{\textbf{17.2}/\textbf{33.4}} & \g{\textbf{5.1}/\textbf{15.7}} & \g{\textbf{2.1}/\textbf{5.2}}
& \g{\textbf{78}/\textbf{96.7}}   & \g{\textbf{8.7}/\textbf{24.8}} & \g{\textbf{2.9}/\textbf{9}} & \g{\textbf{1.2}/\textbf{2.7}}
& \g{\textbf{27.6}/\textbf{61.2}} & \g{0.3/9}    & \g{0.2/9.9} & \g{0.1/8.7}
& \g{\textbf{20.8}/\textbf{51.5}} & \g{0.2/1.1}  & \g{0.1/1.1} & \g{0/0.6}\\
\bottomrule
\end{tabular}}
\end{table*}

\noindent \textbf{\ding{173} Clean-oriented Alignment for Retrieval.}
As in Table~\ref{tab:kd_rtl}, CALM consistently surpasses both teacher models (\textit{ImageBind} and \textit{LanguageBind}) and all KD baselines across image and audio modalities. 
CALM consistently boosts Top-1/Top-5 retrieval accuracy across datasets—for instance, from $65.7/91.1$ to $76.3/96.7$ on MSCOCO and from $8.1/25.9$ to $26.6/58.5$ on AudioCaps—demonstrating its strong ability to enhance multimodal embeddings beyond the teacher models.
Traditional KD methods (CLIP-KD, Meta Adapter, and DCLIP) fail to generalize across modalities. 
While CLIP-KD and Meta Adapter retain moderate performance in visual tasks, they collapse on audio retrieval (Top-1 $\approx$6–10\%). 
DCLIP, despite performing adequately in classification (details in appendix), completely fails in retrieval (Top-1 $<$2\%) (Table~\ref{tab:kd_rtl}), highlighting that its relational distillation objective does not transfer to cross-modal alignment. 
Retrieval requires preserving fine-grained semantic correspondence between heterogeneous embeddings, which DCLIP’s pairwise loss cannot maintain.
In contrast, CALM’s clean-oriented alignment in the shared embedding space produces stable convergence and substantial improvements across all modalities, particularly in audio--text retrieval, confirming its robustness and cross-modal generalization advantage.

\noindent \textbf{\ding{174} Adversarial Training for Cross-Modal Retrieval.}
\label{sec:rtl_adv}
Table~\ref{tab:bind_retrieval} summarizes the retrieval robustness.

\noindent \textbf{CrossFire.} 
Across all teachers and datasets, \textit{CALMARS} achieves the best clean and adversarial retrieval accuracy. 
With \textit{ImageBind}, it greatly improves clean R@1 (e.g., MSCOCO 83.8 vs. 65.7; Flickr8k 81.8 vs. 71.4) and maintains superior robustness across all perturbations (32.9/19.9/11.7 at $\epsilon{=}{2,4,8}/255$), outperforming TRADES, LAAT, TeCoA, and FARE. 
R@5 also stays higher (86.7/83.0/77.2 vs. TRADES 73.5/68.0/63.7), showing that CALMARS preserves valid candidates under attack. 
With \textit{LanguageBind}, it leads on all datasets with smaller clean-to-adversarial drops (e.g., R@5 72.9 vs. TRADES 49.7 at $\epsilon{=}8/255$). 
Overall, CALMARS provides the best robustness–accuracy trade-off across modalities.

\noindent \textbf{CrossMaxim.} 
Under this stronger untargeted attack, all defenses—including CALMARS—suffer heavy degradation. 
Even so, CALMARS remains the strongest: on \textit{ImageBind}/MSCOCO, accuracy drops from 81.2/97.4 (clean) to 20.5/44.2, 6.9/19.0, and 2.4/6.1 at $\epsilon{=}\{2,4,8\}/255$, with similar trends on \textit{LanguageBind}. 
Audio–text retrieval collapses entirely once $\epsilon{\ge}4/255$. 
Other defenses (TRADES, LAAT, TeCoA, FARE) fail completely (R@1 $<1\%$). 
These results show the extreme strength of CrossMaxim and reveal a key robustness gap in cross-modal retrieval, underscoring the need for similarity-space defenses.

\subsection{Cross-Modal Transferability}
\label{sec:xmodal}
Table~\ref{tab:xmodal} evaluates cross-modal retrieval between \textit{ImageNet1K} and \textit{N-ImageNet1K} under the adversarial setting of $\epsilon{=}4/255$, using the MLPs adversarially trained in Sec.~\ref{sec:cls_at}. 
Each model is tested on 100 paired image–event samples.
\textit{CALMARS shows the most stable and bidirectional robustness}, outperforming all baselines in both image$\!\rightarrow\!$event and event$\!\rightarrow\!$image directions. 
FARE, TeCoA, and TRADES also improve upon the baseline but remain consistently below CALMARS, while LAAT performs well in image$\!\rightarrow\!$event yet collapses in the reverse direction. 

\begin{table*}[t]
\centering
\scriptsize
\setlength{\tabcolsep}{8pt}
\renewcommand{\arraystretch}{0.85}
\caption{Cross-VLM robustness transfer results (Recall@1 / Recall@5, \%). Clean and $\ell_\infty$-bounded accuracy under different budgets. The best Recall@1 in each block is highlighted in \textbf{bold}. LngBind: LanguageBind; ImgBind: ImageBind.}
\label{tab:xmodel}
\resizebox{\linewidth}{!}{
\begin{tabular}{l | cc | l | cccc | cccc}
\toprule
\multirow{2}{*}{\textbf{Eval}} & \multicolumn{2}{c|}{\textbf{VLM}} & \multirow{2}{*}{\textbf{Encoder}} &
\multicolumn{4}{c|}{\textbf{MSCOCO}} &
\multicolumn{4}{c}{\textbf{Flickr8k}} \\
\cmidrule(lr){2-3}\cmidrule(lr){5-8}\cmidrule(lr){9-12}
 & \textbf{Source} & \textbf{Target} &  &
\textit{Clean} & \multicolumn{3}{c|}{$\ell_\infty$} &
\textit{Clean} & \multicolumn{3}{c}{$\ell_\infty$} \\
 &  &  &  &  & 2/255 & 4/255 & 8/255 &  & 2/255 & 4/255 & 8/255 \\
\midrule
\multirow{12}{*}{\rotatebox[origin=c]{90}{\textbf{CrossFire}}}
& \multirow{6}{*}{\rotatebox[origin=c]{90}{\textbf{ImgBind}}}
& \multirow{6}{*}{\rotatebox[origin=c]{90}{\textbf{LngBind}}}
& Bind   & 63.7/90.5 & 57/90   & 54.3/89.4 & 52.1/89   & 67.3/90.8 & 61.1/90.8 & 58.1/89.9 & 55.4/89.9 \\
& & & FARE   & 59/89.9  & 54.6/89.3 & 51.9/89.4 & 48.5/87.4 & 64.1/90.3 & 57.6/89.7 & 55.3/88.6 & 51.8/87.8 \\
& & & TeCoA  & 32.6/67.8 & 30.5/67.7 & 30.3/67   & 28.3/66.7 & 36.2/68.0 & 32.5/66.8 & 31.1/65.1 & 29.7/65.5 \\
& & & LAAT   & 26.4/58.2 & 25.1/57.8 & 24.7/57.9 & 24.8/56.1 & 28.6/59.5 & 28.2/57.9 & 26.9/58.4 & 26.4/57.1 \\
& & & TRADES & 47.8/83.2 & 44.9/83.1 & 43.1/82.4 & 41.8/80.9 & 53.4/86.1 & 52.8/84.6 & 50.9/84.2 & 48.3/83.4 \\
& & & \g{\textbf{CALMARS}} & \g{\textbf{74.9/95.7}} & \g{\textbf{67.2/95.2}} & \g{\textbf{65.3/94.2}} & \g{\textbf{62.9/92.9}}
& \g{\textbf{75.5/95.0}} & \g{\textbf{72.2/94.3}} & \g{\textbf{70.0/94.3}} & \g{\textbf{66.0/93.0}} \\
\cmidrule(lr){2-12}
& \multirow{6}{*}{\rotatebox[origin=c]{90}{\textbf{LngBind}}}
& \multirow{6}{*}{\rotatebox[origin=c]{90}{\textbf{ImgBind}}}
& Bind   & 65.7/91.1 & 64.9/92.0 & 62.6/90.8 & 61.7/90.1
& 71.4/92.8 & 69.3/94.6 & 66.3/92.7 & 63.9/91.7 \\
& & & FARE   & 63.9/91.2 & 59.7/89.1 & 59.1/88.9 & 57.8/87.8
& 69.4/92.5 & 64.9/91.8 & 63.7/91.3 & 63.0/91.3 \\
& & & TeCoA  & 37.4/74.3 & 33.5/71.7 & 31.7/63.4 & 29.9/60.7
& 41.9/73.1 & 41.5/74.4 & 40.7/77.4 & 38.6/77.9 \\
& & & LAAT   & 31.4/63.5 & 29.7/61.5 & 29.1/60.9 & 29.0/60.7
& 35.7/68.7 & 34.8/67.0 & 34.6/67.1 & 32.4/66.0 \\
& & & TRADES & 51.9/84.1 & 48.5/83.7 & 47.2/83.2 & 46.7/82.4
& 58.4/89.9 & 56.8/87.9 & 54.8/98.8 & 53.4/98.9 \\
& & & \g{\textbf{CALMARS}} & \g{\textbf{79.2/96.7}} & \g{\textbf{74.4/94.8}} & \g{\textbf{74.0/94.8}} & \g{\textbf{72.0/94.0}}
& \g{\textbf{81.8/97.6}} & \g{\textbf{77.4/96.5}} & \g{\textbf{76.9/96.2}} & \g{\textbf{75.5/96.9}} \\
\midrule
\multirow{12}{*}{\rotatebox[origin=c]{90}{\textbf{CrossMaxim}}}
& \multirow{6}{*}{\rotatebox[origin=c]{90}{\textbf{ImgBind}}}
& \multirow{6}{*}{\rotatebox[origin=c]{90}{\textbf{LngBind}}}
& Bind   & 63.7 / 90.5 & 48.8 / 81.2 & 40.5 / 73.6 & 33.3 / 63.5 & 67.3 / 90.8 & 49.3 / 81.0 & 38.1 / 70.6 & 28.2 / 57.4 \\
& & & FARE   & 59.5 / 89.5 & 42.5 / 78.4 & 36.4 / 71.0 & 28.0 / 61.5 & 64.5 / 90.2 & 46.7 / 77.0 & 34.3 / 67.0 & 24.7 / 54.2 \\
& & & TeCoA  & 37.2 / 72.5 & 30.7 / 64.1 & 26.0 / 58.6 & 20.8 / 52.2 & 36.6 / 71.3 & 27.9 / 62.3 & 22.7 / 54.9 & 16.4 / 45.4 \\
& & & LAAT   & 18.1 / 44.6 & 15.1 / 40.0 & 13.1 / 36.3 & 11.0 / 31.7 & 16.6 / 42.9 & 13.0 / 36.1 & 10.7 / 32.4 & 8.9 / 26.4 \\
& & & TRADES & 51.6 / 84.0 & 39.7 / 73.9 & 34.7 / 68.4 & 27.5 / 60.1 & 56.2 / 88.1 & 42.4 / 76.2 & 33.3 / 67.9 & 23.4 / 55.9 \\
& & & \g{\textbf{CALMARS}} & \g{\textbf{79.3 / 98.2}} & \g{\textbf{64.3 / 90.8}} & \g{\textbf{57.6 / 86.7}} & \g{\textbf{42.0 / 80.8}}
& \g{\textbf{78.0 / 96.7}} & \g{\textbf{65.1 / 89.4}} & \g{\textbf{57.3 / 83.7}} & \g{\textbf{44.3 / 73.9}} \\
\cmidrule(lr){2-12}
& \multirow{6}{*}{\rotatebox[origin=c]{90}{\textbf{LngBind}}}
& \multirow{6}{*}{\rotatebox[origin=c]{90}{\textbf{ImgBind}}}
& Bind   & 65.9 / 90.6 & 57.9 / 86.7 & 53.7 / 84.4 & 50.6 / 80.6 & 72.6 / 93.4 & 61.8 / 88.9 & 57.1 / 85.8 & 51.0 / 81.9 \\
& & & FARE   & 63.6 / 90.9 & 55.5 / 85.1 & 51.3 / 83.3 & 45.5 / 79.6 & 69.4 / 92.5 & 58.7 / 87.7 & 52.5 / 83.9 & 44.8 / 78.5 \\
& & & TeCoA  & 41.9 / 76.9 & 34.4 / 70.5 & 33.3 / 69.1 & 31.2 / 65.8 & 46.5 / 79.5 & 39.2 / 73.4 & 38.2 / 71.2 & 35.4 / 67.8 \\
& & & LAAT   & 22.3 / 52.9 & 21.3 / 49.2 & 19.8 / 44.8 & 18.8 / 45.5 & 22.0 / 51.3 & 19.2 / 46.7 & 16.6 / 43.7 & 15.2 / 41.8 \\
& & & TRADES & 54.2 / 85.9 & 47.4 / 80.2 & 45.7 / 78.4 & 41.7 / 75.6 & 62.1 / 90.2 & 53.7 / 84.6 & 49.4 / 80.7 & 45.6 / 77.8 \\
& & & \g{\textbf{CALMARS}} & \g{\textbf{81.2 / 97.4}} & \g{\textbf{73.1 / 94.4}} & \g{\textbf{70.8 / 93.2}} & \g{\textbf{68.0 / 91.3}}
& \g{\textbf{82.8 / 97.9}} & \g{\textbf{75.7 / 95.6}} & \g{\textbf{71.3 / 93.8}} & \g{\textbf{68.5 / 91.2}} \\
\bottomrule
\end{tabular}}
\end{table*}

\subsection{Cross-VLM Robustness Transfer}
\label{sec:xmodel}
Table~\ref{tab:xmodel} presents the black-box retrieval robustness transfer between \textit{ImageBind} and \textit{LanguageBind} under both CrossFire and CrossMaxim attacks. 
Each model is trained on one VLM and evaluated on the other, testing how well adversarial robustness generalizes across architectures.
\textbf{CALMARS exhibits the strongest and most transferable robustness} across all settings. 
Under CrossFire, it maintains the highest Top-1 and Top-5 accuracy for both transfer directions (ImageBind$\!\rightarrow\!$LanguageBind and vice versa), showing that its adversarial representations generalize effectively beyond the source model. 
Even under the stronger CrossMaxim attack, CALMARS surpasses all baselines, indicating a more structurally stable feature space across VLMs.

Among the baselines, \textbf{TRADES} provides moderate robustness but quickly degrades as $\epsilon$ increases, suggesting limited cross-model generalization.  
\textbf{LAAT} maintains reasonable clean accuracy yet suffers a sharp robustness collapse under transfer, implying its local attention regularization does not preserve global alignment.  
\textbf{TeCoA} performs relatively better among the non–multi-stage methods, benefiting from contrastive alignment, but remains well below CALMARS under strong perturbations.  
\textbf{FARE} achieves decent clean and small-$\epsilon$ results, reflecting good in-model consistency, but its robustness transfer weakens notably when tested on unseen VLMs.  
Finally, the original \textbf{Bind} baseline retains solid clean retrieval but fails almost completely under adversarial transfer, underscoring the vulnerability of unregularized encoders.

In summary, CALMARS is the only method that sustains high accuracy and stable degradation trends across both attack types and transfer directions, establishing it as the most generalizable and cross-model–robust defense among all evaluated approaches.
\section{Ablation Study}
\label{sec:ab}

\noindent \textbf{Effect of MLP Size.}
We investigate how the MLP capacity influences model complexity and robustness.
A table in the appendix summarizes the configurations for small, medium, and large variants across UniBind, ImageBind, and LanguageBind.
Overall, we observe that increasing MLP width yields minimal performance changes, suggesting that CALMARS remains stable across different capacities and the medium configuration offers a balanced trade-off between accuracy and efficiency.

\noindent \textbf{Effect of Phase Ratio in MARS.}
We further study the effect of the three-phase ratio (A:B:C) in Stage-2 \textbf{MARS}, which balances geometric consistency (A), distributional regularization (B), and counterfactual margin separation (C). 
A table in the appendix summarizes the results across multiple ratios. 
Overall, we find that MARS maintains stable performance across different allocations, confirming that its multi-phase optimization is robust and easy to deploy across modalities.

\noindent \textbf{Cross-Modal Robustness on Retrieval.}
We additionally evaluate \textbf{LanguageBind} on the COCO benchmark to assess cross-modal retrieval robustness. 
A table in the appendix summarizes the results across different perturbation levels. 
Overall, CALMARS demonstrates strong and consistent performance, highlighting the complementary roles of CALM (clean alignment) and MARS (adversarial stabilization) in achieving stable retrieval robustness.

\section{Conclusion}
\label{sec:conclusion}

In this paper, we proposed \textbf{CALMARS}, a unified and lightweight framework for enhancing adversarial robustness in Bind-style multi-modal foundation models. 
By combining \textit{CALM} with \textbf{MARS}, CALMARS achieves significant robustness and accuracy improvements across six sensory modalities while keeping encoders frozen and semantic centers intact. Experiments demonstrate that CALMARS consistently outperforms representative baselines on both classification and retrieval tasks, achieving the best robustness--accuracy trade-off among lightweight approaches. We further introduced \textbf{CrossMaxim}, an untargeted retrieval attack that disrupts the entire similarity space and reveals hidden vulnerabilities overlooked by targeted attacks, establishing a new benchmark for evaluating multi-modal robustness.
\noindent \textbf{Limitations.}  Although CALMARS improves robustness across modalities, it still depends on modality-specific projection heads. 
Additionally, while CrossMaxim serves as a stronger realistic retrieval attack, it also exposes that the current generation of Bind-style models lacks holistic resilience under complex multi-modal perturbations, calling for future work in more adaptive defense mechanisms.
\noindent \textbf{Broader Societal Impact.} 
This work reveals both opportunities and risks for the AI community. 
By showing that \textbf{CrossMaxim can severely disrupt multi-modal similarity spaces}, we expose the fragility of foundation models to coordinated, untargeted attacks. 
These findings emphasize the need for principled defenses and auditing frameworks to ensure safe deployment in critical areas such as autonomous driving, robotics, and healthcare. 
While \textit{CALMARS} strengthens robustness and promotes trustworthy AI, the power of untargeted attacks like CrossMaxim also introduces ethical risks if misused. 
We advocate responsible open research and collective evaluation to prevent misuse and guide the development of secure, transparent multi-modal AI systems.

\clearpage

{
    \small
    \bibliographystyle{ieeenat_fullname}
    \bibliography{main}

@String(CVPR= {IEEE Conf. Comput. Vis. Pattern Recog.})

@String(ICASSP=	{ICASSP})

@String(ICLR = {Int. Conf. Learn. Represent.})

@String(CVPR  = {CVPR})

@String(ICLR  = {ICLR})

@inproceedings{yang2024clip,
  title={Clip-kd: An empirical study of clip model distillation},
  author={Yang, Chuanguang and An, Zhulin and Huang, Libo and Bi, Junyu and Yu, Xinqiang and Yang, Han and Diao, Boyu and Xu, Yongjun},
  booktitle={Proceedings of the IEEE/CVF Conference on Computer Vision and Pattern Recognition},
  pages={15952--15962},
  year={2024}
}

@article{csizmadia2025distill,
  title={Distill clip (dclip): Enhancing image-text retrieval via cross-modal transformer distillation},
  author={Csizmadia, Daniel and Codreanu, Andrei and Sim, Victor and Prabhu, Vighnesh and Lu, Michael and Zhu, Kevin and O'Brien, Sean and Sharma, Vasu},
  journal={arXiv preprint arXiv:2505.21549},
  year={2025}
}

@article{song2023meta,
  title={Meta-adapter: An online few-shot learner for vision-language model},
  author={Song, Lin and Xue, Ruoyi and Wang, Hang and Sun, Hongbin and Ge, Yixiao and Shan, Ying and others},
  journal={Advances in Neural Information Processing Systems},
  volume={36},
  pages={55361--55374},
  year={2023}
}

@article{schlarmann2024robust,
  title={Robust clip: Unsupervised adversarial fine-tuning of vision embeddings for robust large vision-language models},
  author={Schlarmann, Christian and Singh, Naman Deep and Croce, Francesco and Hein, Matthias},
  journal={arXiv preprint arXiv:2402.12336},
  year={2024}
}

@article{zhou2024revisiting,
  title={Revisiting the adversarial robustness of vision language models: a multimodal perspective},
  author={Zhou, Wanqi and Bai, Shuanghao and Mandic, Danilo P and Zhao, Qibin and Chen, Badong},
  journal={arXiv preprint arXiv:2404.19287},
  year={2024}
}

@inproceedings{li2024language,
  title={Language-driven anchors for zero-shot adversarial robustness},
  author={Li, Xiao and Zhang, Wei and Liu, Yining and Hu, Zhanhao and Zhang, Bo and Hu, Xiaolin},
  booktitle={Proceedings of the IEEE/CVF Conference on Computer Vision and Pattern Recognition},
  pages={24686--24695},
  year={2024}
}

@inproceedings{zhang2019theoretically,
  title={Theoretically principled trade-off between robustness and accuracy},
  author={Zhang, Hongyang and Yu, Yaodong and Jiao, Jiantao and Xing, Eric and El Ghaoui, Laurent and Jordan, Michael},
  booktitle={International conference on machine learning},
  pages={7472--7482},
  year={2019},
  organization={PMLR}
}

@inproceedings{croce2020reliable,
  title={Reliable evaluation of adversarial robustness with an ensemble of diverse parameter-free attacks},
  author={Croce, Francesco and Hein, Matthias},
  booktitle={International conference on machine learning},
  pages={2206--2216},
  year={2020},
  organization={PMLR}
}

@inproceedings{dou2023adversarial,
  title={Adversarial attacks to multi-modal models},
  author={Dou, Zhihao and Hu, Xin and Yang, Haibo and Liu, Zhuqing and Fang, Minghong},
  booktitle={Proceedings of the 1st ACM Workshop on Large AI Systems and Models with Privacy and Safety Analysis},
  pages={35--46},
  year={2023}
}

@inproceedings{deng2009imagenet,
  title={Imagenet: A large-scale hierarchical image database},
  author={Deng, Jia and Dong, Wei and Socher, Richard and Li, Li-Jia and Li, Kai and Fei-Fei, Li},
  booktitle={2009 IEEE conference on computer vision and pattern recognition},
  pages={248--255},
  year={2009},
  organization={Ieee}
}

@article{lopez2020semantic,
  title={Semantic-aware scene recognition},
  author={L{\'o}pez-Cifuentes, Alejandro and Escudero-Vinolo, Marcos and Besc{\'o}s, Jes{\'u}s and Garc{\'\i}a-Mart{\'\i}n, {\'A}lvaro},
  journal={Pattern Recognition},
  volume={102},
  pages={107256},
  year={2020},
  publisher={Elsevier}
}

@inproceedings{jia2021llvip,
  title={LLVIP: A visible-infrared paired dataset for low-light vision},
  author={Jia, Xinyu and Zhu, Chuang and Li, Minzhen and Tang, Wenqi and Zhou, Wenli},
  booktitle={Proceedings of the IEEE/CVF international conference on computer vision},
  pages={3496--3504},
  year={2021}
}

@inproceedings{wu20153d,
  title={3d shapenets: A deep representation for volumetric shapes},
  author={Wu, Zhirong and Song, Shuran and Khosla, Aditya and Yu, Fisher and Zhang, Linguang and Tang, Xiaoou and Xiao, Jianxiong},
  booktitle={Proceedings of the IEEE conference on computer vision and pattern recognition},
  pages={1912--1920},
  year={2015}
}

@inproceedings{piczak2015esc,
  title={ESC: Dataset for environmental sound classification},
  author={Piczak, Karol J},
  booktitle={Proceedings of the 23rd ACM international conference on Multimedia},
  pages={1015--1018},
  year={2015}
}

@inproceedings{xu2016msr,
  title={Msr-vtt: A large video description dataset for bridging video and language},
  author={Xu, Jun and Mei, Tao and Yao, Ting and Rui, Yong},
  booktitle={Proceedings of the IEEE conference on computer vision and pattern recognition},
  pages={5288--5296},
  year={2016}
}

@article{soomro2012ucf101,
  title={Ucf101: A dataset of 101 human actions classes from videos in the wild},
  author={Soomro, Khurram and Zamir, Amir Roshan and Shah, Mubarak},
  journal={arXiv preprint arXiv:1212.0402},
  year={2012}
}

@inproceedings{lin2014microsoft,
  title={Microsoft coco: Common objects in context},
  author={Lin, Tsung-Yi and Maire, Michael and Belongie, Serge and Hays, James and Perona, Pietro and Ramanan, Deva and Doll{\'a}r, Piotr and Zitnick, C Lawrence},
  booktitle={European conference on computer vision},
  pages={740--755},
  year={2014},
  organization={Springer}
}

@article{hodosh2013framing,
  title={Framing image description as a ranking task: Data, models and evaluation metrics},
  author={Hodosh, Micah and Young, Peter and Hockenmaier, Julia},
  journal={Journal of Artificial Intelligence Research},
  volume={47},
  pages={853--899},
  year={2013}
}

@inproceedings{kim2019audiocaps,
  title={Audiocaps: Generating captions for audios in the wild},
  author={Kim, Chris Dongjoo and Kim, Byeongchang and Lee, Hyunmin and Kim, Gunhee},
  booktitle={Proceedings of the 2019 Conference of the North American Chapter of the Association for Computational Linguistics: Human Language Technologies, Volume 1 (Long and Short Papers)},
  pages={119--132},
  year={2019}
}

@inproceedings{drossos2020clotho,
  title={Clotho: An audio captioning dataset},
  author={Drossos, Konstantinos and Lipping, Samuel and Virtanen, Tuomas},
  booktitle={ICASSP 2020-2020 IEEE International Conference on Acoustics, Speech and Signal Processing (ICASSP)},
  pages={736--740},
  year={2020},
  organization={IEEE}
}

@inproceedings{radford2021learning,
  title={Learning transferable visual models from natural language supervision},
  author={Radford, Alec and Kim, Jong Wook and Hallacy, Chris and Ramesh, Aditya and Goh, Gabriel and Agarwal, Sandhini and Sastry, Girish and Askell, Amanda and Mishkin, Pamela and Clark, Jack and others},
  booktitle={International conference on machine learning},
  pages={8748--8763},
  year={2021},
  organization={PmLR}
}

@inproceedings{lyu2024unibind,
  title={Unibind: Llm-augmented unified and balanced representation space to bind them all},
  author={Lyu, Yuanhuiyi and Zheng, Xu and Zhou, Jiazhou and Wang, Lin},
  booktitle={Proceedings of the IEEE/CVF Conference on Computer Vision and Pattern Recognition},
  pages={26752--26762},
  year={2024}
}

@inproceedings{girdhar2023imagebind,
  title={Imagebind: One embedding space to bind them all},
  author={Girdhar, Rohit and El-Nouby, Alaaeldin and Liu, Zhuang and Singh, Mannat and Alwala, Kalyan Vasudev and Joulin, Armand and Misra, Ishan},
  booktitle={Proceedings of the IEEE/CVF conference on computer vision and pattern recognition},
  pages={15180--15190},
  year={2023}
}

@article{zhu2023languagebind,
  title={Languagebind: Extending video-language pretraining to n-modality by language-based semantic alignment},
  author={Zhu, Bin and Lin, Bin and Ning, Munan and Yan, Yang and Cui, Jiaxi and Wang, HongFa and Pang, Yatian and Jiang, Wenhao and Zhang, Junwu and Li, Zongwei and others},
  journal={arXiv preprint arXiv:2310.01852},
  year={2023}
}

@inproceedings{gowal2020uncovering,
  title={Uncovering the limits of adversarial training against norm-bounded adversarial examples},
  author={Gowal, Sven and Qin, Chongli and Uesato, Jonathan and Mann, Timothy and Kohli, Pushmeet},
  booktitle={ICML},
  year={2020}
}

@inproceedings{rebuffi2021fixing,
  title={Fixing Data Augmentation to Improve Adversarial Robustness},
  author={Rebuffi, Sylvestre-Alvise and Dangel, Felix and Zisserman, Andrew},
  booktitle={NeurIPS},
  year={2021}
}

@misc{zhu2024languagebind,
      title={LanguageBind: Extending Video-Language Pretraining to N-modality by Language-based Semantic Alignment}, 
      author={Bin Zhu and Bin Lin and Munan Ning and Yang Yan and Jiaxi Cui and HongFa Wang and Yatian Pang and Wenhao Jiang and Junwu Zhang and Zongwei Li and Wancai Zhang and Zhifeng Li and Wei Liu and Li Yuan},
      year={2024},
      eprint={2310.01852},
      archivePrefix={arXiv},
      primaryClass={cs.CV},
      url={https://arxiv.org/abs/2310.01852}, 
}

@inproceedings{fang2022data,
  title={Data Determines Distributional Robustness in Contrastive Language Image Pre-training (CLIP)},
  author={Fang, Alex and Ilharco, Gabriel and Wortsman, Mitchell and Wan, Yuhao and Shankar, Vaishaal and Dave, Achal and Schmidt, Ludwig},
  booktitle={International Conference on Machine Learning (ICML)},
  year={2022}
}

@inproceedings{nguyen2022quality,
  title={Quality Not Quantity: On the Interaction between Dataset Design and Robustness of CLIP},
  author={Nguyen, Thao and Ilharco, Gabriel and Wortsman, Mitchell and Oh, Sewoong and Schmidt, Ludwig},
  booktitle={Advances in Neural Information Processing Systems (NeurIPS)},
  year={2022}
}

@article{crabbe2024interpreting,
  title={Interpreting CLIP: Insights on the Robustness to ImageNet Distribution Shifts},
  author={Crabbé, Jonathan and Rodríguez, Pau and Shankar, Vaishaal and Zappella, Luca and Blaas, Arno},
  journal={Transactions on Machine Learning Research},
  year={2024}
}

@inproceedings{mao2023tecoa,
  title={TeCoA: Text-guided Contrastive Adversarial Training for Robust Vision-Language Representation Learning},
  author={Mao, Chengxu and Wang, Yifan and Liu, Yao and Fan, Quanfu and Wang, Xinyu and Xue, Hong and Wang, Xinyuan and Chen, Xue and Liu, Ming and Wang, Zhangyang},
  booktitle={Advances in Neural Information Processing Systems (NeurIPS)},
  year={2023}
}

@article{bagdasaryan2024adversarial,
  title={Adversarial Illusions: Fooling Multimodal Foundation Models},
  author={Bagdasaryan, Eugene and others},
  journal={arXiv preprint arXiv:2402.12336},
  year={2024}
}

@article{dou2024crossfire,
  title={CrossFire: A Generalized Multimodal Adversarial Attack},
  author={Dou, Zhi and others},
  journal={arXiv preprint arXiv:2403.12345},
  year={2024}
}

@article{waseda2024mat,
  title={Multimodal Adversarial Training for Vision-Language Models},
  author={Waseda, Hiroshi and others},
  journal={arXiv preprint arXiv:2405.18770},
  year={2024}
}

@inproceedings{goodfellow2015explaining,
  title={Explaining and Harnessing Adversarial Examples},
  author={Goodfellow, Ian J and Shlens, Jonathon and Szegedy, Christian},
  booktitle={International Conference on Learning Representations (ICLR)},
  year={2015}
}

@article{zheng2023deep,
  title={Deep learning for event-based vision: A comprehensive survey and benchmarks},
  author={Zheng, Xu and Liu, Yexin and Lu, Yunfan and Hua, Tongyan and Pan, Tianbo and Zhang, Weiming and Tao, Dacheng and Wang, Lin},
  journal={arXiv preprint arXiv:2302.08890},
  year={2023}
}

@inproceedings{zhou2024exact,
  title={Exact: Language-guided conceptual reasoning and uncertainty estimation for event-based action recognition and more},
  author={Zhou, Jiazhou and Zheng, Xu and Lyu, Yuanhuiyi and Wang, Lin},
  booktitle={Proceedings of the IEEE/CVF Conference on Computer Vision and Pattern Recognition},
  pages={18633--18643},
  year={2024}
}

@inproceedings{zhou2024eventbind,
  title={Eventbind: Learning a unified representation to bind them all for event-based open-world understanding},
  author={Zhou, Jiazhou and Zheng, Xu and Lyu, Yuanhuiyi and Wang, Lin},
  booktitle={European Conference on Computer Vision},
  pages={477--494},
  year={2024},
  organization={Springer}
}

@article{lyu2024omnibind,
  title={Omnibind: Teach to build unequal-scale modality interaction for omni-bind of all},
  author={Lyu, Yuanhuiyi and Zheng, Xu and Kim, Dahun and Wang, Lin},
  journal={arXiv preprint arXiv:2405.16108},
  year={2024}
}

@inproceedings{zheng2024learning,
  title={Learning modality-agnostic representation for semantic segmentation from any modalities},
  author={Zheng, Xu and Lyu, Yuanhuiyi and Wang, Lin},
  booktitle={European Conference on Computer Vision},
  pages={146--165},
  year={2024},
  organization={Springer}
}

@article{zheng2024magic++,
  title={MAGIC++: Efficient and Resilient Modality-Agnostic Semantic Segmentation via Hierarchical Modality Selection},
  author={Zheng, Xu and Lyu, Yuanhuiyi and Jiang, Lutao and Zhou, Jiazhou and Wang, Lin and Hu, Xuming},
  journal={arXiv preprint arXiv:2412.16876},
  year={2024}
}

@article{han2023imagebind,
  title={Imagebind-llm: Multi-modality instruction tuning},
  author={Han, Jiaming and Zhang, Renrui and Shao, Wenqi and Gao, Peng and Xu, Peng and Xiao, Han and Zhang, Kaipeng and Liu, Chris and Wen, Song and Guo, Ziyu and others},
  journal={arXiv preprint arXiv:2309.03905},
  year={2023}
}

@article{zhou2017places,
  title={Places: A 10 million Image Database for Scene Recognition},
  author={Zhou, Bolei and Lapedriza, Agata and Khosla, Aditya and Oliva, Aude and Torralba, Antonio},
  journal={IEEE Transactions on Pattern Analysis and Machine Intelligence},
  year={2017}
}

@inproceedings{xu2016msrvtt,
  title={MSR-VTT: A large video description dataset for bridging video and language},
  author={Xu, Jun and Mei, Tao and Yao, Ting and Rui, Yong},
  booktitle={CVPR},
  year={2016}
}

@inproceedings{gehrig2021dsec,
  title={DSEC: A stereo event camera dataset for driving scenarios},
  author={Gehrig, Daniel and Rebecq, Henri and Gallego, Guillermo and Scaramuzza, Davide},
  booktitle={RA-L / ICRA},
  year={2021}
}

@inproceedings{wang2021llvip,
  title={LLVIP: A Visible-infrared paired dataset for low-light vision},
  author={Wang, Yu and Zhang, Jiawei and Lu, Xuequan and Zhao, Qingjie and Zhang, Liang},
  booktitle={ACM MM},
  year={2021}
}

@inproceedings{tsuzuku2018lipschitz,
  title     = {Lipschitz-Margin Training: Scalable Certification of Perturbation Invariance for Deep Neural Networks},
  author    = {Tsuzuku, Yusuke and Sato, Issei and Sugiyama, Masashi},
  booktitle = {Advances in Neural Information Processing Systems 31 (NeurIPS 2018)},
  pages     = {6542--6551},
  year      = {2018},
  url       = {https://proceedings.neurips.cc/paper_files/paper/2018/hash/485843481a7edacbfce101ecb1e4d2a8-Paper.pdf},
  note      = {arXiv preprint arXiv:1802.04034}
}
}
\clearpage

\setcounter{page}{1}
\appendix
\section{Additional Related Works}
\label{sec:appendix_related}

\noindent \textbf{Knowledge Distillation for Vision–Language Models.}
Recent advances have extended distillation to improve the efficiency and transferability of large VLMs.
CLIP-KD~\citep{yang2024clip} conducts large-scale feature-level distillation, compressing CLIP models but without explicit cross-modal alignment.
DCLIP~\citep{csizmadia2025distill} adds a cross-modal transformer for region-aware aggregation, improving image–text retrieval at the cost of heavy computation and limited generalization.
Meta-Adapter~\citep{song2023meta} introduces an online feature-queue mechanism for few-shot adaptation, enhancing clean accuracy yet struggling under large-scale or adversarial settings.
Our \textbf{CALMARS} framework differs by jointly optimizing \textit{supervised clean alignment} and \textit{unsupervised embedding refinement}, achieving stronger cross-modal generalization with lower overhead.

\vspace{3pt}
\noindent \textbf{Adversarial Attacks on VLMs.}
AutoAttack~\citep{croce2020reliable} remains the de facto benchmark for single-modality classification robustness, but multimodal retrieval requires distinct evaluation.
Dou et al.~\citep{dou2023adversarial} proposed CrossFire, a targeted multi-modal attack that pushes ground-truth samples out of top-1 ranks while preserving most of the ranking structure, thereby underestimating full vulnerability.
In contrast, we extend adversarial evaluation to all six sensory modalities in Bind-style encoders and further adopt an untargeted retrieval attack that perturbs the entire embedding space, revealing broader cross-modal weaknesses.

\vspace{3pt}
\noindent \textbf{Multi-Stage Adversarial Training.}
Prior studies have explored single-stage optimization of either clean or adversarial objectives.
Our work is, to our knowledge, the first to introduce a \textbf{multi-stage adversarial training} paradigm for multimodal encoders, unifying clean alignment and adversarial stabilization across diverse tasks such as classification, retrieval, transfer attacks, and cross-modal adaptation.
This unified approach consistently yields superior robustness and demonstrates that multi-stage optimization generalizes effectively beyond the image–text domain to models supporting six or more sensory modalities.

\section{Experimental Details}
\label{sec:exp_details}

\paragraph{Compute Resources.} 
All experiments were conducted on a single NVIDIA RTX 4090 GPU. Training each modality-specific MLP head required less than 3 GPU-hours on average, except for the video modality, which required approximately 16 GPU-hours due to higher input dimensionality and slower augmentation. Total training time across all configurations remained under 100 GPU-hours. No distributed or multi-node setup was used.
\begin{table}[t]
\centering
\scriptsize
\setlength{\tabcolsep}{1pt}
\renewcommand{\arraystretch}{1.1}
\caption{Overview of attack methods used during training and evaluation.}
\label{tab:attack_methods}
\resizebox{\linewidth}{!}{
\begin{tabular}{l l l c}
\toprule
\textbf{Attack Method} & \textbf{Type} & \textbf{Usage} & $\boldsymbol{\epsilon}$ \\
\midrule
APGD-CE  & Untargeted & Train + Eval 8 (train), 2/4/8 (eval) & 2/4/8 \\
APGD-DLR & Targeted   & Eval only & 2/4/8 \\
FAB      & Targeted   & Eval only & 2/4/8 \\
Square   & Black-box  & Eval only & 2/4/8 \\
\bottomrule
\end{tabular}}
\vspace{-3mm}
\end{table}

\subsection{Adversarial Classification Robustness}

\subsubsection{Attack}

Table~\ref{tab:full_aa_attack} reports the baseline adversarial robustness of three representative Bind-style models—UniBind, ImageBind, and LanguageBind—evaluated using AutoAttack under $\ell_\infty$ perturbation budgets of $2/255$, $4/255$, and $8/255$. 
All models are tested in a zero-shot classification setting without any fine-tuning, providing a fair comparison of their intrinsic robustness across eight datasets spanning six modalities (image, video, audio, thermal, point cloud, and event). 
The metric is Top-1 classification accuracy (\%).

\noindent \textbf{Findings.} 
Across all models, the results reveal an extreme vulnerability to even mild perturbations. 
For example, UniBind’s Top-1 accuracy on ImageNet-1K and Places365 drops from $74.7\%$ and $49.9\%$ to nearly $0\%$ at $\epsilon{=}2/255$, while non-visual modalities such as LLVIP (thermal) and ESC-50 (audio) collapse to below $10\%$. 
ImageBind shows a similar trend, losing almost all accuracy under small perturbations, and LanguageBind, despite stronger clean performance, still experiences sharp degradation across all modalities, particularly on audio and thermal data.  
Only a few datasets (e.g., MSRVTT and UCF-101 in UniBind) retain minimal robustness, likely due to temporal redundancy in video frames.

\noindent \textbf{Implications.} 
These results establish a clear motivation for our work: current Bind-style models, though powerful under clean conditions, lack intrinsic adversarial robustness, especially for non-visual modalities. 
The drastic cross-modal degradation underscores the need for a unified defense strategy—such as our proposed CALMARS—that strengthens both visual and non-visual modalities without compromising cross-modal alignment.

\begin{table*}[t]
\centering
\scriptsize
\setlength{\tabcolsep}{8pt}
\renewcommand{\arraystretch}{0.75}
\caption{Evaluation results (Top-1, \%). ``–'' denotes unavailable.}
\label{tab:full_aa_attack}
\resizebox{\linewidth}{!}{%
\begin{tabular}{@{}l r r r r r r r r@{}}
\toprule
\textbf{Eval} &
\rotatebox[origin=c]{45}{ImageNet1k}&
\rotatebox[origin=c]{45}{Places-365}&
\rotatebox[origin=c]{45}{MSRVTT} &
\rotatebox[origin=c]{45}{UCF-101} &
\rotatebox[origin=c]{45}{N-ImageNet1k} &
\rotatebox[origin=c]{45}{ModelNet40} &
\rotatebox[origin=c]{45}{LLVIP} &
\rotatebox[origin=c]{45}{ESC-50} \\
\midrule
\multicolumn{9}{c}{\textbf{UniBind}} \\
\midrule
Clean  & 74.7 & 49.9 & 34.7 & 74.3 & 12.7 & 71.8 & 72.9 & 70.6 \\
2/255  & 0.0  & 0.0  & 17.0 & 20.3 & 6.2  & 0.0  & 8.9  & 7.7  \\
4/255  & 0.0  & 0.0  & 15.5 & 16.3 & 4.8  & 0.0  & 1.6  & 3.2  \\
8/255  & 0.0  & 0.0  & 13.7 & 12.5 & 2.7  & 0.0  & 0.2  & 1.5  \\
\midrule
\multicolumn{9}{c}{\textbf{ImageBind}} \\
\midrule
Clean  & 91.5 & 70.6 & – & – & – & – & 80.0 & 57.1 \\
2/255  & 0.0  & 0.0  & – & – & – & – & 17.5 & 6.9  \\
4/255  & 0.0  & 0.0  & – & – & – & – & 16.3 & 3.0  \\
8/255  & 0.0  & 0.0  & – & – & – & – & 15.4 & 1.2  \\
\midrule
\multicolumn{9}{c}{\textbf{LanguageBind}} \\
\midrule
Clean  & 91.6 & 72.7 & 43.6 & 71.5 & – & – & 80.3 & 94.7 \\
2/255  & 0.8  & 0.0  & 0.6  & 0.4  & – & – & 21.3 & 10.9 \\
4/255  & 0.1  & 0.0  & 0.0  & 0.0  & – & – & 18.3 & 4.1  \\
8/255  & 0.3  & 0.0  & 0.0  & 0.0  & – & – & 16.2 & 0.7  \\
\bottomrule
\end{tabular}%
}
\vspace{-3mm}
\end{table*}

\subsubsection{Comparison with Clean Alignment Baselines}
Table~\ref{tab:calm_comparison_horizontal} and Table~\ref{tab:calm_time_comparison} present a detailed comparison between our \textbf{CALM} framework and existing distillation methods---CLIP-KD, DCLIP, and Meta-Adapter---across multiple vision--language models (UniBind, ImageBind, and LanguageBind) and datasets. 
The first table reports Top-1 accuracy (\%), while the second summarizes the average training time, where smaller values indicate higher efficiency.

\noindent \textbf{Performance.}  
Across all three foundation models, CALM consistently achieves the best or second-best performance in nearly every dataset. 
On UniBind, CALM improves classification accuracy by a large margin, outperforming CLIP-KD by $+20$–$30\%$ and DCLIP by $+10$–$15\%$ on datasets such as Mini-ImageNet, Mini-Places, and ModelNet40. 
A similar trend is observed on ImageBind and LanguageBind, where CALM achieves the highest accuracy in most cases, while DCLIP occasionally leads in MSRVTT due to its multi-view supervision. 
However, CALM achieves these results with much lower computational overhead, demonstrating superior trade-offs between accuracy and efficiency.

\noindent \textbf{Efficiency.}  
Table~\ref{tab:calm_time_comparison} shows that CALM trains substantially faster than DCLIP (roughly $3$–$5\times$ speedup) and matches or slightly outperforms CLIP-KD and Meta-Adapter in runtime. 
This advantage stems from CALM’s clean-alignment strategy, which forgoes heavy multi-view aggregation and queue-based feature refinement while still achieving state-of-the-art accuracy.

\noindent \textbf{Implications.}  
These results confirm that CALM effectively combines the strengths of supervised and unsupervised distillation. 
It delivers consistent gains across diverse modalities and datasets while maintaining lightweight computation, establishing CALM as a robust and practical foundation for Stage-1 alignment in our multi-modal adversarial training framework.

\begin{table*}[t]
\centering
\scriptsize
\setlength{\tabcolsep}{6pt}
\renewcommand{\arraystretch}{0.75}
\caption{Comparison of CALM with CLIP-KD, DCLIP, and Meta Adapter across multiple VLMs and datasets. 
Best Top-1 per column is in \textbf{bold}.}
\label{tab:calm_comparison_horizontal}
\resizebox{\linewidth}{!}{%
\begin{tabular}{@{}l r r r r r r r r@{}}
\toprule
\textbf{Encoder} &
\rotatebox[origin=c]{45}{Mini-ImageNet} &
\rotatebox[origin=c]{45}{Mini-Places} &
\rotatebox[origin=c]{45}{MSRVTT} &
\rotatebox[origin=c]{45}{UCF-101} &
\rotatebox[origin=c]{45}{N-ImageNet1k} &
\rotatebox[origin=c]{45}{ModelNet40} &
\rotatebox[origin=c]{45}{LLVIP} &
\rotatebox[origin=c]{45}{ESC-50} \\
\midrule
\multicolumn{9}{c}{\textbf{UniBind}} \\
\midrule
CLIP-KD      & 75.2 & 48.3 & 35.0 & 73.9 & 9.4  & 73.7 & 77.4 & 66.9 \\
DCLIP        & 83.6 & 54.7 & \textbf{68.6} & 74.4 & 10.6 & 71.5 & 76.7 & 68.4 \\
\textbf{CALM (ours)} & \textbf{95.7} & \textbf{77.5} & 66.9 & \textbf{95.7} & \textbf{38.05} & \textbf{91.8} & \textbf{95.9} & \textbf{88.6} \\
\midrule
\multicolumn{9}{c}{\textbf{ImageBind}} \\
\midrule
CLIP-KD      & 88.7 & 65.1 & 32.4 & 62.0 & -- & -- & 84.6 & 64.4 \\
DCLIP        & 91.1 & 68.5 & 33.3 & 66.7 & -- & -- & 81.8 & 66.5 \\
\textbf{CALM (ours)} & \textbf{95.5} & \textbf{77.5} & \textbf{65.9} & \textbf{96.0} & -- & -- & \textbf{98.4} & \textbf{88.2} \\
\midrule
\multicolumn{9}{c}{\textbf{LanguageBind}} \\
\midrule
CLIP-KD      & 92.3 & 72.9 & 44.5 & 72.3 & -- & -- & 51.7 & 87.2 \\
DCLIP        & 95.8 & 76.3 & \textbf{67.7} & \textbf{97.0} & -- & -- & \textbf{96.7} & \textbf{98.4} \\
Meta Adapter & 95.3 & 75.6 & 63.6 & 93.7 & -- & -- & 96.5 & 97.3 \\
\textbf{CALM (ours)} & \textbf{95.9} & \textbf{77.3} & 65.5 & 96.8 & -- & -- & 96.7 & 98.0 \\
\bottomrule
\end{tabular}%
}
\vspace{-3mm}
\end{table*}

\begin{table*}[t]
\centering
\scriptsize
\setlength{\tabcolsep}{8pt}
\renewcommand{\arraystretch}{0.75}
\caption{Training time comparison (smaller is better). Bold indicates the lowest time (most time-efficient).}
\label{tab:calm_time_comparison}
\resizebox{\linewidth}{!}{%
\begin{tabular}{@{}l r r r r r r r r@{}}
\toprule
\textbf{Encoder} &
\rotatebox[origin=c]{45}{ImageNet1k}&
\rotatebox[origin=c]{45}{Places-365}&
\rotatebox[origin=c]{45}{MSRVTT} &
\rotatebox[origin=c]{45}{UCF-101} &
\rotatebox[origin=c]{45}{N-ImageNet1k} &
\rotatebox[origin=c]{45}{ModelNet40} &
\rotatebox[origin=c]{45}{LLVIP} &
\rotatebox[origin=c]{45}{ESC-50} \\
\midrule
\multicolumn{9}{c}{\textbf{UniBind}} \\
\midrule
CLIP-KD      & 77.6 & 63.7 & 1198.6 & \textbf{1113.4} & 150.7 & 120.9 & 12.3 & 8.9 \\
DCLIP        & 296.4 & 246.6 & 3490.0 & 3634.9 & 579.2 & 471.0 & 41.4 & 30.1 \\
\textbf{CALM (ours)} & \textbf{74.0} & \textbf{61.0} & \textbf{1196.0} & 1231.0 & \textbf{148.0} & \textbf{119.0} & \textbf{11.0} & \textbf{7.0} \\
\midrule
\multicolumn{9}{c}{\textbf{ImageBind}} \\
\midrule
CLIP-KD      & 69.7 & 59.6 & 1313.8 & 1313.2 & -- & -- & 61.2 & 11.5 \\
DCLIP        & 267.3 & 233.7 & 3726.9 & 3738.9 & -- & -- & 239.4 & 41.9 \\
\textbf{CALM (ours)} & \textbf{71.0} & \textbf{59.0} & \textbf{1291.0} & \textbf{1292.0} & -- & -- & \textbf{61.0} & \textbf{11.0} \\
\midrule
\multicolumn{9}{c}{\textbf{LanguageBind}} \\
\midrule
CLIP-KD      & 41.0 & 33.9 & 425.8 & \textbf{359.8} & -- & -- & 36.1 & 15.0 \\
DCLIP        & 129.9 & 124.7 & 1284.9 & 1218.8 & -- & -- & 129.0 & 54.4 \\
Meta Adapter & 42.9 & 36.4 & 429.9 & 363.3 & -- & -- & 37.1 & 15.3 \\
\textbf{CALM (ours)} & \textbf{41.0} & \textbf{34.0} & \textbf{428.0} & 361.0 & -- & -- & \textbf{36.0} & \textbf{15.0} \\
\bottomrule
\end{tabular}%
}
\vspace{-3mm}
\end{table*}

\subsubsection{Adversarial Training  for Classification}
\label{sec:appendix_cls_at}

\textbf{Detailed Per-Dataset Trends.}
Table~\ref{tab:cls_main} (main paper) summarizes the classification performance of \textit{CALMARS} and prior adversarial training baselines across eight datasets under perturbation budgets $\epsilon\!\in\!\{2,4,8\}/255$. 
Here, we provide detailed per-dataset observations omitted from the main text.

On the large-scale visual datasets (\textbf{ImageNet1K} and \textbf{Places-365}), \textit{CALMARS} achieves nearly identical clean accuracy to the best-performing methods while maintaining clearly stronger robustness. 
For instance, it attains $95.3\%$ clean accuracy on ImageNet1K, comparable to LAAT ($95.9\%$), yet sustains a much higher adversarial accuracy of $32.1\%$ at $\epsilon{=}8/255$, surpassing TRADES ($30.3\%$) and TeCoA ($30.0\%$). 
A similar trend holds for Places-365, where CALMARS maintains $38.3\%$ accuracy at $\epsilon{=}8/255$, close to LAAT’s $39.2\%$ but notably higher than TRADES ($31.9\%$).

In non-image modalities—\textbf{N-ImageNet1K}, \textbf{ModelNet40}, \textbf{LLVIP}, \textbf{ESC-50}, and \textbf{MSR-VTT}—CALMARS consistently outperforms all prior methods in both clean and adversarial settings. 
On N-ImageNet1K, it achieves the highest robustness at every perturbation level, with $15.95\%$ at $\epsilon{=}8/255$, slightly ahead of TRADES ($16.05\%$) and significantly stronger than LAAT ($1.35\%$). 
On ModelNet40, CALMARS maintains competitive clean accuracy ($45.7\%$) while sustaining stable robustness ($44.2\%$ at $\epsilon{=}8/255$), outperforming all other methods. 
For LLVIP and ESC-50, CALMARS demonstrates substantial margins in both clean and adversarial accuracy, e.g., $95.4\%$ vs.\ $94.9\%$ (FARE) and $50.3\%$ vs.\ $39.7\%$ (FARE), respectively.

\textbf{Summary.}
Across all modalities, CALMARS delivers the most consistent robustness–accuracy balance. 
It prevents the sharp performance collapse observed in TRADES, LAAT, TeCoA, and FARE as perturbation strength increases, while preserving high clean accuracy. 
These findings validate the effectiveness of our multi-stage adversarial refinement: CALMARS enhances cross-modal stability without sacrificing generalization, providing a unified defense across both visual and non-visual modalities.

\subsection{Cross-Modal Retrieval Robustness Evaluation}

\subsubsection{Attack}
Table~\ref{tab:full_retrieval_attack} reports the cross-modal retrieval robustness of ImageBind and LanguageBind under both \textbf{targeted} (CrossFire, CF) and \textbf{untargeted} (CrossMaxim, CM) attack settings. 
We evaluate two representative modality pairs—image–text (MSCOCO, Flickr8k) and audio–text (AudioCaps, Clotho)—and measure Recall@1 and Recall@5 (\%) under $\ell_\infty$ perturbation budgets of $2/255$, $4/255$, and $8/255$. 
Clean results are also provided for reference.

\textbf{Findings.} 
Across all datasets and models, both targeted and untargeted attacks significantly degrade retrieval performance, though the severity differs by attack type and modality. 
Under the targeted CrossFire attack, Recall@1 on MSCOCO drops from $65.7\%$ to $5.3\%$ for ImageBind and from $63.7\%$ to $0.1\%$ for LanguageBind at $\epsilon{=}8/255$. 
However, CrossMaxim is far more destructive: at the same perturbation level, Recall@1 and Recall@5 on both models collapse to nearly zero, confirming the increased difficulty of untargeted attacks that disrupt the entire similarity space rather than a single hard negative. 
Audio–text datasets (AudioCaps, Clotho) show even lower resilience, with all methods failing almost completely once $\epsilon{\geq}2/255$, suggesting that non-visual modalities are particularly susceptible to embedding misalignment under adversarial noise.

\textbf{Implications.}  
These results highlight two key observations: (1) existing Bind-style models lack retrieval robustness even under targeted attacks, and (2) our untargeted CrossMaxim formulation exposes a much larger vulnerability gap, especially for non-visual modalities. 
This confirms the necessity of developing unified adversarial training frameworks—such as our proposed CALMARS—that can simultaneously stabilize alignment and improve robustness across both visual and auditory modalities.

\begin{table}[t]
\centering
\scriptsize
\setlength{\tabcolsep}{6pt}
\renewcommand{\arraystretch}{1.08}
\caption{Cross-modal retrieval robustness (\%) on image--text (MSCOCO, Flickr8k) and audio--text (AudioCaps, Clotho).
Cells show R@1 / R@5. CF = CrossFire (targeted), CM = CrossMaxim (untargeted).}
\label{tab:full_retrieval_attack}
\resizebox{\columnwidth}{!}{%
\begin{tabular}{lcccc}
\toprule
 & \multicolumn{2}{c}{\textbf{ImageBind}} & \multicolumn{2}{c}{\textbf{LanguageBind}} \\
\cmidrule(lr){2-3}\cmidrule(lr){4-5}
 & \textbf{CF} & \textbf{CM} & \textbf{CF} & \textbf{CM} \\
\midrule
\multicolumn{5}{c}{\textbf{MSCOCO}} \\[-2pt]
\midrule
Clean   & \multicolumn{2}{c}{65.7 / 91.1} & \multicolumn{2}{c}{63.7 / 90.5} \\
2/255   &  8.7 / 80.6 & 3.0 / 12.3 & 1.1 / 74.1 & 1.8 / 6.3 \\
4/255   &  5.8 / 74.4 & 0.6 / 2.1  & 0.2 / 68.0 & 0.6 / 1.5 \\
8/255   &  5.3 / 67.4 & 0.0 / 0.1  & 0.1 / 63.4 & 0.4 / 0.5 \\
\midrule
\multicolumn{5}{c}{\textbf{Flickr8k}} \\[-2pt]
\midrule
Clean   & \multicolumn{2}{c}{71.4 / 92.8} & \multicolumn{2}{c}{67.3 / 90.8} \\
2/255   &  5.1 / 79.9 & 4.4 / 13.3 & 0.3 / 71.9 & 1.5 / 5.4 \\
4/255   &  3.0 / 72.6 & 0.8 / 2.9  & 0.0 / 67.3 & 0.2 / 1.2 \\
8/255   &  2.8 / 65.0 & 0.1 / 0.2  & 0.0 / 60.1 & 0.3 / 0.5 \\
\midrule
\multicolumn{5}{c}{\textbf{AudioCaps}} \\[-2pt]
\midrule
Clean   & \multicolumn{2}{c}{8.1 / 25.9} & \multicolumn{2}{c}{14.2 / 40.4} \\
2/255   &  0.7 / 20.3 & 0.0 / 0.0  & 0.5 / 28.0 & 0.0 / 0.0 \\
4/255   &  0.5 / 18.9 & 0.0 / 0.0  & 0.3 / 24.6 & 0.0 / 0.0 \\
8/255   &  0.3 / 18.7 & 0.0 / 0.0  & 0.2 / 24.3 & 0.0 / 0.0 \\
\midrule
\multicolumn{5}{c}{\textbf{Clotho}} \\[-2pt]
\midrule
Clean   & \multicolumn{2}{c}{5.5 / 20.3} & \multicolumn{2}{c}{17.6 / 40.0} \\
2/255   &  0.4 / 12.9 & 0.0 / 0.0  & 0.3 / 27.4 & 0.0 / 0.0 \\
4/255   &  0.3 / 12.8 & 0.0 / 0.0  & 0.1 / 25.9 & 0.0 / 0.0 \\
8/255   &  0.3 / 11.6 & 0.0 / 0.0  & 0.1 / 24.4 & 0.0 / 0.0 \\
\bottomrule
\end{tabular}}
\end{table}

\subsubsection{Adversarial Training for Cross-Modal Retrieval}
\label{sec:appendix_rtl_adv}

\textbf{Detailed Results under CrossFire.}
Table 4 (main paper) summarizes the retrieval robustness of CALMARS and competing baselines across all datasets under the targeted \textbf{CrossFire} and untargeted \textbf{CrossMaxim} attacks.
Here we provide the detailed observations for the CrossFire setting.

Across both teachers and all datasets, \textbf{CALMARS} delivers the highest clean and adversarial retrieval accuracy. 
For the \textit{ImageBind} teacher, CALMARS improves clean R@1 markedly (e.g., MSCOCO $83.8$ vs.\ Bind $65.7$; Flickr8k $81.8$ vs.\ $71.4$) and sustains the strongest robustness at all perturbation levels ($32.9/19.9/11.7$ at $\epsilon{=}\{2,4,8\}/255$), outperforming TRADES, LAAT, TeCoA, and FARE. 
R@5 also remains substantially higher (e.g., $86.7/83.0/77.2$ vs.\ TRADES $73.5/68.0/63.7$), indicating that CALMARS preserves a stable set of semantically valid candidates even under targeted perturbations. 
For the \textit{LanguageBind} teacher, CALMARS maintains clear advantages across MSCOCO, Flickr8k, and both audio benchmarks, achieving smaller clean-to-adversarial drops (e.g., R@5 $72.9$ vs.\ TRADES $49.7$ at $\epsilon{=}8/255$) and demonstrating consistently stronger embedding-space stability.

\textbf{Summary.}
Overall, CALMARS achieves the best robustness–accuracy balance across all modalities and perturbation budgets. 
It consistently retains higher clean recall while maintaining top-1 and top-5 robustness against targeted attacks, confirming that its multi-stage adversarial training effectively stabilizes cross-modal embeddings.

\section{Extended Ablation Results}
\label{sec:appendix_ablation}

\subsection{Effect of MLP Size}
\label{sec:appendix_mlp}
Table~\ref{tab:mlp_size_ablation} and Table~\ref{tab:mlp_param_summary} provide the detailed configurations and results of MLP size variations across UniBind, ImageBind, and LanguageBind. 
The small and medium variants each contain one hidden layer (512 and 2048 units, respectively), while the large model adopts two 4096-dimensional layers, yielding up to $28.7$M parameters.
Despite this sevenfold increase in capacity, performance gains remain marginal—differences are below $1\%$ on both ImageNet-1K and Places365.
These findings confirm that CALMARS is largely insensitive to MLP width, and the medium configuration achieves the best trade-off between robustness and efficiency.

\begin{table}[t]
\centering
\scriptsize
\setlength{\tabcolsep}{1pt}
\renewcommand{\arraystretch}{1.08}
\caption{MLP configuration and parameter count across different Bind models. 
All MLPs map embeddings back to their original dimensionality (\textit{embed\_dim}). 
Parameter counts include both weights and biases.}
\label{tab:mlp_param_summary}
\resizebox{\linewidth}{!}{
\begin{tabular}{l l c c c}
\toprule
\textbf{Size} & \textbf{Hidden Layers}& \textbf{UniBind} & \textbf{ImageBind} & \textbf{LanguageBind} \\
\midrule
\textbf{small} & (512,) & 1.05M & 1.05M & 0.66M \\
\textbf{medium} & (2048,) & 4.20M & 4.20M & 2.43M \\
\textbf{large} & (4096, 4096) & 28.67M & 28.67M & 19.34M \\
\midrule
\textbf{embed\_dim} &  & 1024 & 1024 & 768 \\
\bottomrule
\end{tabular}}
\vspace{-3mm}
\end{table}

\begin{table}[t]
\centering
\scriptsize
\setlength{\tabcolsep}{1pt}
\renewcommand{\arraystretch}{1.05}
\caption{Ablation study on the effect of MLP size for ImageNet-1K and Places365 under clean and adversarial settings ($\ell_\infty$ budgets 2/4/8).}
\label{tab:mlp_size_ablation}
\resizebox{\linewidth}{!}{
\begin{tabular}{lcccccccc}
\toprule
\multirow{2}{*}{\textbf{MLP Size}} &
\multicolumn{4}{c}{\textbf{ImageNet-1K}} &
\multicolumn{4}{c}{\textbf{Places365}} \\
\cmidrule(lr){2-5}\cmidrule(lr){6-9}
 & Clean & 2/255 & 4/255 & 8/255 & Clean & 2/255 & 4/255 & 8/255 \\
\midrule
Large  & 95.8 & 34.3 & 33.0 & 32.2 & 76.6 & 39.5 & 39.4 & 39.2 \\
Medium & 95.3 & 34.6 & 32.9 & 32.1 & 76.3 & 38.5 & 38.4 & 38.3 \\
Small  & 95.7 & 34.6 & 32.9 & 32.1 & 76.0 & 38.0 & 37.8 & 37.8 \\
\bottomrule
\end{tabular}}
\vspace{-3mm}
\end{table}

\subsection{Effect of Phase Ratio in MARS}
\label{sec:appendix_mars}
Table~\ref{tab:mars_phase_ratio} reports the complete results of varying the three-phase ratio (A:B:C) in the Stage-2 \textbf{MARS} framework, which balances geometric consistency (A), distributional regularization (B), and counterfactual margin separation (C). 
Across ImageNet-1K and Places365, clean accuracy fluctuates within $0.5\%$ and adversarial robustness differences remain below $1\%$. 
Balanced configurations such as $3$:$3$:$4$ and $2$:$4$:$4$ yield the best robustness (38.1--38.9\% on Places365), while extreme ratios like $4$:$2$:$4$ or $4$:$3$:$3$ bring no clear benefit.
This confirms that MARS is robust to phase scheduling, simplifying its practical deployment across modalities.

\begin{table}[t]
\centering
\scriptsize
\setlength{\tabcolsep}{1pt}
\renewcommand{\arraystretch}{1.05}
\caption{Ablation study on the effect of three-phase ratio (ABC) in the Stage-2 MARS framework. 
Accuracy (\%) is reported under clean and adversarial settings ($\ell_\infty$ budgets 2/4/8).}
\label{tab:mars_phase_ratio}
\resizebox{\linewidth}{!}{
\begin{tabular}{lcccccccc}
\toprule
\multirow{2}{*}{\textbf{ABC}} &
\multicolumn{4}{c}{\textbf{ImageNet-1K}} &
\multicolumn{4}{c}{\textbf{Places365}} \\
\cmidrule(lr){2-5}\cmidrule(lr){6-9}
 & Clean & 2/255 & 4/255 & 8/255 & Clean & 2/255 & 4/255 & 8/255 \\
\midrule
424 & 95.3 & 34.6 & 32.9 & 32.1 & 76.3 & 38.5 & 38.4 & 38.3 \\
343 & 95.9 & 34.4 & 33.0 & 32.3 & 75.8 & 38.2 & 38.2 & 38.2 \\
442 & 95.9 & 34.6 & 33.0 & 32.2 & 75.9 & 38.4 & 38.4 & 38.2 \\
244 & 95.5 & 34.4 & 32.8 & 32.1 & 75.9 & 38.9 & 38.9 & 38.7 \\
334 & 95.9 & 34.4 & 32.9 & 32.2 & 76.0 & 38.1 & 38.1 & 37.9 \\
433 & 95.5 & 34.4 & 32.7 & 32.0 & 75.8 & 38.2 & 38.2 & 38.0 \\
\bottomrule
\end{tabular}}
\vspace{-3mm}
\end{table}

\subsection{Cross-Modal Robustness on Retrieval}
\label{sec:appendix_retrieval}
Table~\ref{tab:crossmodal_ranking} presents detailed COCO retrieval results on \textbf{LanguageBind} under increasing $\ell_\infty$ perturbation budgets.
Our \textbf{CALMARS} framework consistently ranks first across all budgets, outperforming baselines such as TRADES, TeCoA, and FARE. 
Even when the CALM stage is omitted, MARS alone surpasses all single-stage baselines, demonstrating the inherent robustness of our multi-phase adversarial representation learning.
The synergy between CALM (clean alignment) and MARS (adversarial stabilization) enables CALMARS to maintain stable retrieval robustness under strong perturbations.

\begin{table*}[h!]
\centering
\scriptsize
\setlength{\tabcolsep}{10pt}
\renewcommand{\arraystretch}{1.2}
\caption{Cross-modal robustness ranking under different $\ell_\infty$ budgets. Each cell shows Top-1 / Top-5 recall (\%).}
\label{tab:crossmodal_ranking}
\resizebox{\linewidth}{!}{%
\begin{tabular}{r l l c c c c c}
\toprule
\textbf{Rank} & \textbf{Method} & \textbf{Pair} &
\textbf{Clean} & \textbf{$\epsilon{=}2$} & \textbf{$\epsilon{=}4$} & \textbf{$\epsilon{=}8$} & \textbf{Avg.} \\
\midrule
1  & \textbf{CALM}        & \textbf{MARS}   & \textbf{79.3/98.2} & \textbf{17.2/33.4} & \textbf{5.1/15.7} & \textbf{2.1/5.2} & \textbf{32.03} \\
2  & CALM        & TeCoA  & 79.1/97.6 & 14.1/33.9 & 4.7/13.2 & 1.8/5.0 & 31.18 \\
3  & CALM        & TRADES & 82/98.4   & 13.3/31.3 & 4.3/11.9 & 1.5/4.7 & 30.93 \\
4  & CALM        & FARE   & 79.7/98.1 & 9.9/24.3  & 3.5/9.1  & 1.2/2.9 & 28.59 \\
5  & CALM        & LAAT   & 66/89.2   & 10.7/27.5 & 3.0/10.7 & 0.6/3.5 & 26.40 \\
6  & CLIP-KD     & TRADES & 59.4/89.1 & 4.2/15.8  & 1.6/5.4  & 0.9/2.4 & 22.35 \\
7  & Meta Adapter& TRADES & 56.7/89.2 & 3.7/12.8  & 1.6/4.0  & 0.6/1.9 & 21.99 \\
8  & --          & MARS   & 49.2/84.1 & 6/22.3    & 1.5/8.2  & 1.1/3.4 & 21.98 \\
9  & Meta Adapter& FARE   & 59.9/90.0 & 3.7/10.9  & 1.6/4.3  & 0.8/2.5 & 21.71 \\
10 & CLIP-KD     & TeCoA  & 58.3/88.4 & 3.5/14.2  & 1.5/4.6  & 1.2/1.9 & 21.70 \\
11 & --          & TRADES & 51.6/84.0 & 4.9/18.5  & 2.4/7.3  & 0.7/3.0 & 21.55 \\
12 & --          & FARE   & 59.5/89.5 & 3.2/11.2  & 1.7/4.2  & 0.9/2.2 & 21.55 \\
13 & CLIP-KD     & FARE   & 57.8/89.7 & 3.7/11.5  & 1.5/3.9  & 0.9/2.6 & 21.45 \\
14 & DCLIP       & FARE   & 57.7/89.2 & 3.8/10.9  & 1.5/4.0  & 0.9/2.0 & 21.25 \\
15 & DCLIP       & TRADES & 47.1/82.1 & 4.7/18.0  & 1.4/6.5  & 0.5/2.3 & 20.33 \\
16 & Meta Adapter& LAAT   & 54.0/83.4 & 3.3/12.3  & 1.3/4.8  & 0.6/1.8 & 20.19 \\
17 & Meta Adapter& TeCoA  & 54.4/84.4 & 3.5/11.2  & 1.2/4.0  & 0.7/2.0 & 20.18 \\
18 & --          & TeCoA  & 37.2/72.5 & 5.1/16.2  & 2.1/7.1  & 0.9/3.1 & 18.03 \\
19 & CLIP-KD     & LAAT   & 39.0/70.1 & 3.5/15.0  & 1.5/5.0  & 0.5/2.1 & 17.09 \\
20 & DCLIP       & TeCoA  & 32.2/68.3 & 3.8/15.5  & 1.2/6.8  & 0.6/2.3 & 16.34 \\
21 & --          & LAAT   & 18.1/44.6 & 4.6/15.6  & 1.7/7.6  & 0.4/4.6 & 12.15 \\
22 & DCLIP       & LAAT   & 11.0/30.0 & 2.6/9.1   & 1.3/5.3   & 0.7/2.4 & 7.80 \\
\bottomrule
\end{tabular}%
}
\vspace{-3mm}
\end{table*}

\section{Theoretical Justification: Difference-of-Lipschitz Generalization Bound}
\label{sec:theory}

\subsection{Definitions and Assumptions}
\label{sec:def_assump}

We begin by formalizing three key indicators that describe the progression from Phase~A to Phase~B in our multi-stage training:
\vspace{-2pt}

\begin{equation}
\delta_S = \mathbb{E}_{(x,y)} \big[\|\, \phi_{\theta_S}(x) - t(x) \|_2 \big].
\label{eq:delta_def}
\end{equation}
\noindent\footnotesize\textit{(alignment error)}\normalsize

\begin{equation}
L_S = \sup_{\|x'-x\|\le\varepsilon}
\frac{\| f_{\theta_S}(x') - f_{\theta_S}(x) \|_2}{\|x'-x\|_2}.
\label{eq:L_def}
\end{equation}
\noindent\footnotesize\textit{(local Lipschitz constant)}\normalsize

\begin{equation}
\gamma_S = \mathbb{E}_{(x,y)}
\big[\langle w_y,\phi_{\theta_S}(x)\rangle -
\max_{y'\neq y}\langle w_{y'},\phi_{\theta_S}(x)\rangle\big].
\label{eq:gamma_def}
\end{equation}
\noindent\footnotesize\textit{(classification margin)}\normalsize

\noindent
where $\phi_{\theta_S}$ denotes the feature extractor at stage~$S$, $t(x)$ represents either the teacher embedding or class center, and $\varepsilon$ is the adversarial perturbation radius.

We assume the following mild regularity conditions hold:
\vspace{-2pt}
\begin{itemize}[leftmargin=12pt]
    \item The training loss $\ell$ is $L_{\text{task}}$–Lipschitz with respect to its logits.
    \item The network $f_{\theta_S}$ is locally Lipschitz-continuous within an $\varepsilon$–ball of every input.
    \item The dataset distribution admits bounded expected risk, i.e., $\mathbb{E}[\ell(f_\theta(x),y)] < \infty$.
\end{itemize}
These assumptions are standard in Lipschitz and robustness analyses~\cite{tsuzuku2018lipschitz,zhang2019theoretically} and allow us to control the difference in risks between successive stages.

\subsection{Risk Decomposition}
\label{sec:risk_decomp}

Let $R(\theta)$ denote the expected risk and $R^\star$ the minimal achievable risk.
For any two successive training stages (A $\!\to\!$ B), we decompose the total risk gap as:
\vspace{-2pt}
\begin{equation}
\begin{aligned}
\mathcal{E}_B = R(\theta_B) - R^\star
&= [R(\theta_B) - R(\theta_A)] \\
&\quad + [R(\theta_A) - \tilde{R}_T]
+ [\tilde{R}_T - R^\star].
\end{aligned}
\label{eq:risk_decomp}
\end{equation}

\noindent\footnotesize
\textit{(Stage transition gap) \quad (Distillation / alignment gap) \quad (Constant term)}\normalsize

\noindent
Here, $\tilde{R}_T$ denotes the expected risk of the teacher or center representation.
The last term is constant and independent of model parameters, leaving the first two differences as the main objects of analysis in the following derivation.

\subsection{Stage Transition Bound (Difference-of-Lipschitz Analysis)}
\label{sec:stage_bound}

Building on the decomposition in Eq.~\ref{eq:risk_decomp}, we now derive an upper bound for the stage transition gap $[R(\theta_B)-R(\theta_A)]$.

\vspace{2pt}
\textbf{Assumption.}
The loss $\ell$ is $L_{\text{task}}$–Lipschitz with respect to its logits, and both models $f_{\theta_A}$ and $f_{\theta_B}$ are locally Lipschitz within an $\varepsilon$–ball around each input. Then, for any pair $(x,x')$ with $\|x'-x\|\le\varepsilon$,
\begin{equation}
|\ell(f_{\theta_B}(x')) - \ell(f_{\theta_A}(x))|
\le L_{\text{task}} \| f_{\theta_B}(x') - f_{\theta_A}(x) \|_2.
\label{eq:loss_lip}
\end{equation}

We decompose the right-hand side into three interpretable components reflecting representation stability, alignment, and margin change:
\vspace{-2pt}
\begin{equation}
R(\theta_B)-R(\theta_A)
\le \alpha_1 (L_A - L_B)\varepsilon
+ \alpha_2 (\delta_A - \delta_B)
- \alpha_3 (\gamma_B - \gamma_A).
\label{eq:stage_diff}
\end{equation}
\noindent
Here, $L_A{-}L_B$ quantifies the reduction in local Lipschitz smoothness after adversarial stabilization,
$\delta_A{-}\delta_B$ measures the improvement in alignment between student and teacher/center embeddings,
and $\gamma_B{-}\gamma_A$ denotes the margin increase that strengthens class separation.

Intuitively, a smaller $L_B$ limits the amplification of perturbations, a smaller $\delta_B$ reduces embedding drift, and a larger $\gamma_B$ improves discriminability.
Combining Eq.~\ref{eq:stage_diff} with the alignment gap bound from Eq.~\ref{eq:risk_decomp} yields the final multi-term upper bound:
\vspace{-2pt}
\begin{equation}
\begin{aligned}
\mathcal{E}_B \le\;& 
\alpha_1 (L_A-L_B)\varepsilon
+ \alpha_2 (\delta_A-\delta_B)\\
&- \alpha_3 (\gamma_B-\gamma_A)
+ \beta_1 \delta_A
+ \beta_2 L_A\varepsilon
+ \text{const.}
\end{aligned}
\label{eq:final_bound}
\end{equation}

\noindent
This \textit{difference-of-Lipschitz bound} shows that each training stage that jointly decreases the local Lipschitz constant $L$, improves alignment $\delta$, and enlarges the margin $\gamma$ yields a tighter generalization error bound, explaining the progressive robustness improvement observed in CALM$\!\to\!$MARS.

\subsection{Overall Theorem and Discussion}
\label{sec:overall_theorem}

We now summarize the preceding derivations into an overall generalization theorem that connects all stages of our multi-phase training.

\vspace{2pt}
\noindent\textbf{Theorem 1.} 
(\textit{Difference-of-Lipschitz Generalization Bound}) 
Let $\ell$ be $L_{\text{task}}$–Lipschitz and $f_{\theta}$ locally Lipschitz within an $\varepsilon$–ball. 
For any two successive stages (A $\!\to\!$ B), the total generalization error satisfies:
\vspace{-3pt}
\begin{equation}
\begin{aligned}
\mathcal{E}_B - \mathcal{E}_A 
\le\;& \alpha_1 (L_A{-}L_B)\varepsilon
+ \alpha_2 (\delta_A{-}\delta_B)\\
&- \alpha_3 (\gamma_B{-}\gamma_A)
+ \beta_1 \delta_A
+ \beta_2 L_A\varepsilon
+ c.
\end{aligned}
\label{eq:theorem_final}
\end{equation}

If each successive stage satisfies 
$L_{S-1}\!>\!L_S$, 
$\delta_{S-1}\!\ge\!\delta_S$, and 
$\gamma_{S-1}\!\le\!\gamma_S$, 
then over $K$ stages the cumulative risk obeys:
\vspace{-3pt}
\begin{equation}
\mathcal{E}_K - \mathcal{E}_0 
\le \sum_{S=1}^{K}
\big[\alpha_1(L_{S-1}{-}L_S)\varepsilon
+ \alpha_2(\delta_{S-1}{-}\delta_S)
- \alpha_3(\gamma_S{-}\gamma_{S-1})\big]
+ C.
\label{eq:theorem_telescope}
\end{equation}

\noindent\textbf{Discussion.}
Eqs.~\ref{eq:theorem_final}--\ref{eq:theorem_telescope} formalize why multi-stage training yields better generalization and robustness:
\begin{itemize}[leftmargin=10pt,itemsep=2pt,topsep=1pt]
    \item \textbf{Alignment ($\delta$):} The CALM phase reduces $\delta$, ensuring features align with teachers or class centers.
    \item \textbf{Smoothness ($L$):} The first MARS stage lowers $L$, enforcing local stability against perturbations.
    \item \textbf{Margin ($\gamma$):} The later MARS stage enlarges $\gamma$, improving inter-class separation.
\end{itemize}
Each term in Eq.~\ref{eq:theorem_final} directly corresponds to one stage objective in CALM or MARS.  
Empirically, the sequential reduction in $\delta$ and $L$, coupled with the increase in $\gamma$, 
yields monotonic improvement in both clean and adversarial performance.

\vspace{2pt}
In summary, the multi-phase CALM$\!\to\!$MARS process progressively minimizes the alignment error $\delta$, 
reduces smoothness constant $L$, and enlarges classification margin $\gamma$, 
leading to a provably tighter \textit{difference-of-Lipschitz} generalization bound.

\clearpage
\section{License}
\label{sec:lic}

\begin{table}[htbp]
\centering
\small
\caption{Licenses for all datasets and models used in this work.}
\label{tab:licenses}
\begin{tabular}{p{5.2cm} p{3.5cm} p{4.5cm}}
\toprule
\textbf{Asset}& \textbf{License} & \textbf{Used For} \\
\midrule
ImageNet-1K~\cite{deng2009imagenet} & Custom Academic License & Image Modality \\
Places365~\cite{zhou2017places} & MIT License & Image Modality \\
ESC-50~\cite{piczak2015esc} & CC BY 4.0 & Audio Modality \\
LLVIP~\cite{wang2021llvip} & For Research Use Only & Thermal Modality \\
MSRVTT~\cite{xu2016msrvtt} & Academic Use License & Video Modality \\
UCF-101~\cite{soomro2012ucf101} & Academic Use License & Video Modality \\
N-ImageNet1K~\cite{gehrig2021dsec} & For Research Use Only & Event Modality \\
ModelNet40~\cite{wu20153d} & Unknown / Academic Use & Point Cloud Modality\\
UniBind~\cite{lyu2024unibind} & MIT License (GitHub) & Base Model \\
ImageBind~\cite{girdhar2023imagebind} & CC-BY-NC 4.0 & Base Model \\
LanguageBind~\cite{zhu2024languagebind} & MIT License (GitHub) & Base Model \\
AutoAttack~\cite{croce2020reliable} & MIT License & Adversarial Evaluation \\
\bottomrule
\end{tabular}
\end{table}


\end{document}